%% file: main.tex
\documentclass[preprint,12pt]{elsarticle}

\usepackage{times}
\usepackage{soul}
\usepackage{url}
\usepackage[hidelinks]{hyperref}
\usepackage[T1]{fontenc}
\usepackage[utf8]{inputenc}
\usepackage[small]{caption}
\usepackage{graphicx}
\usepackage{amsmath}
\usepackage{amsthm}
\usepackage{booktabs}
\usepackage{algorithm}
\usepackage{algorithmic}
\usepackage{array,multirow}
\usepackage{subfigure}
\usepackage{multicol} 
\usepackage{amsfonts}
\usepackage{stackengine}
\usepackage{xcolor}
\usepackage{mathtools}
\usepackage{wrapfig}
\usepackage[export]{adjustbox}

\newcommand\ours{{Adapt \& Align}}

\input{defs}

\journal{Neural Networks}

\begin{document}

\begin{frontmatter}
    


\title{Adapt \& Align: Continual Learning with \\ Generative Models' Latent Space Alignment}

\author[inst1,inst2]{Kamil Deja}
\author[inst1]{Bartosz Cywiński}
\author[inst1]{Jan Rybarczyk}
\author[inst1,inst2,inst3]{Tomasz Trzciński}

\affiliation[inst1]{organization={Warsaw University of Technology},
            addressline={pl. Politechniki 1}, 
            city={Warsaw},
            postcode={00-661}, 
            country={Poland}}

\affiliation[inst2]{organization={NCBR Ideas},
            addressline={ul. Chmielna 69}, 
            city={Warsaw},
            postcode={00-801}, 
            country={Poland}}
            
\affiliation[inst3]{organization={Tooploox},
            addressline={Tęczowa 7}, 
            city={Wrocław},
            postcode={53-601}, 
            country={Poland}}

\begin{abstract}

In this work, we introduce \ours{}, a method for continual learning of neural networks by aligning latent representations in generative models. Neural Networks suffer from abrupt loss in performance when retrained with additional training data from different distributions. At the same time, training with additional data without access to the previous examples rarely improves the model's performance. In this work, we propose a new method that mitigates those problems by employing generative models and splitting the process of their update into two parts. In the first one, we train a \emph{local} generative model using only data from a new task. In the second phase, we consolidate latent representations from the \emph{local} model with a \emph{global} one that encodes knowledge of all past experiences. We introduce our approach with Variational Auteoncoders and Generative Adversarial Networks. Moreover, we show how we can use those generative models as a general method for continual knowledge consolidation that can be used in downstream tasks such as classification.

\end{abstract}




\end{frontmatter}


\section{Introduction}

\begin{figure}[!htbp]
	\centering
	\includegraphics[width=0.85\linewidth]{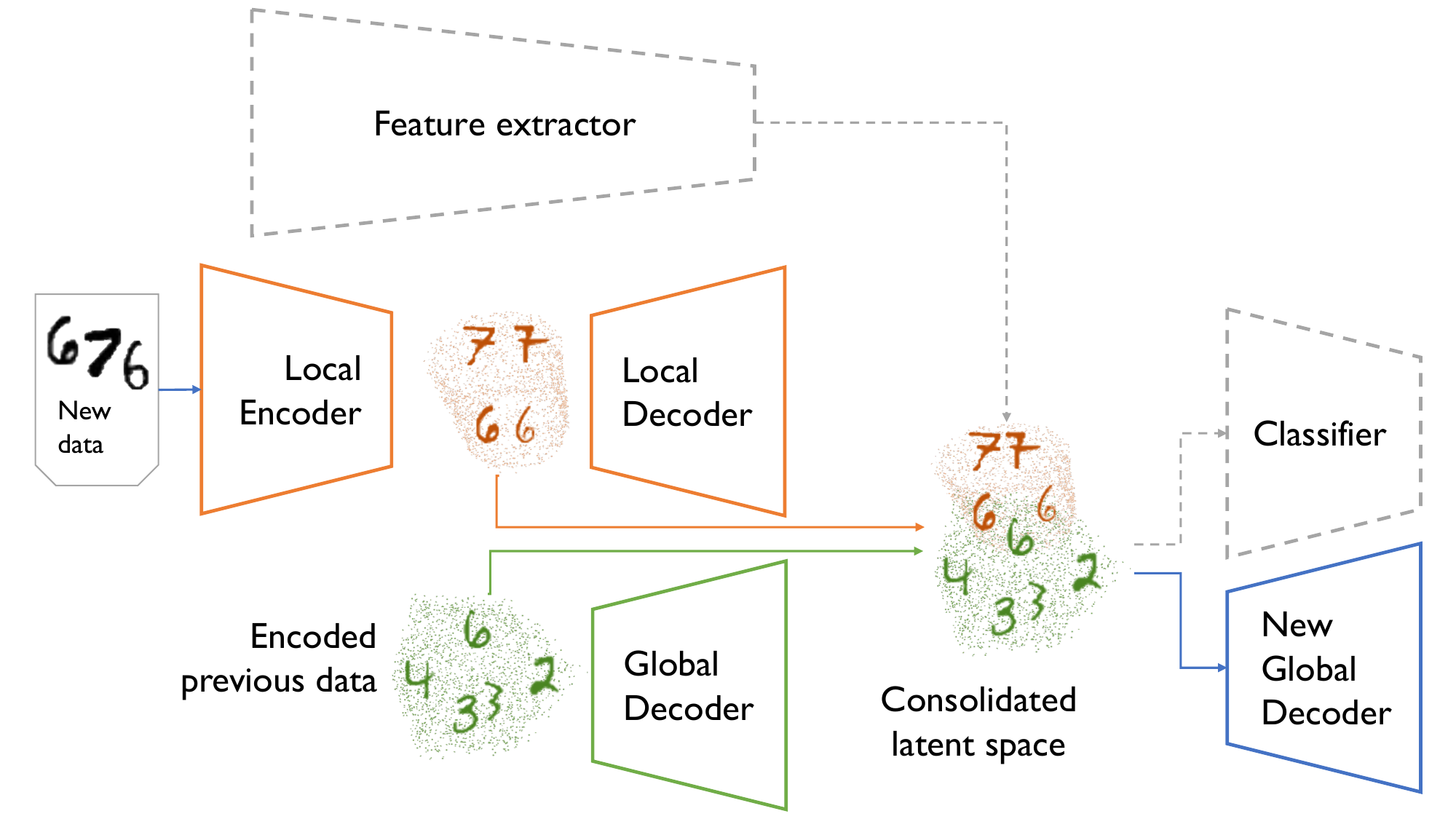}
	\caption{Overview of our \ours{} based on VAE. We first learn a \textit{local} copy of our model with each new task to encode new data examples. Then, we consolidate those with our current global decoder, the main model that is able to generate examples from all tasks. Additionally, we train a classifier aligned with consolidated latent space}
	\label{fig:teaser}
\end{figure}
Recent advances in generative models~\citep{2014goodfellow,kingma2014autoencoding,sohl2015deep} led to their unprecedented proliferation across many real-life applications. 
This includes synthetic images generation~\citep{brock2018large,dhariwal2021diffusion,song2020denoising}, with the extension to the methods creating new images from textual prompts~\citep{ramesh2022hierarchical, saharia22}, speech and music synthesis systems~\cite{popov2021grad, oord2016wavenet, yang2017midinet} or scientific computing such as particles collisions simulations~\citep{paganini2018calogan,deja2020end,kansal2021particle}. 

These applications are possible because generative models effectively represent complex data manifolds. However, this maintaining high-quality representations remains difficult to deliver in real-life situations where training data is presented to the model in separate portions. 
Moreover, data distributions between these portions often vary significantly. Hence, updating the model with new examples in a naive way leads to {\it catastrophic forgetting} of previous knowledge. In discriminative models, this can be observed as a loss in accuracy on classes from the previous tasks, while in generative modeling, catastrophic forgetting manifests through a limited distribution of generated examples. 

\emph{Generative continual learning} methods aim to address this challenge
usually in one of three ways: through regularization~(e.g.~\citep{nguyen2017variational}), adjustment of the structure of a network to the next task ~(e.g.~\citep{rao2019continual}), or rehearsal of previously seen samples when training with new data (e.g.~\citep{rebuffi2017icarl}). 
In our previous work \cite{deja2022multiband}, we observed that all of the recently proposed solutions focus on an artificial task-based class-incremental (CI) scenario where consecutive data portions contain disjoint classes and share almost no similarity. We argued that this assumption simplifies the problem and introduced a new scenario that focuses on the continual consolidation of knowledge from a stream of any tasks, including those sharing some similarities.

\newpage
While, for the CI scenario, we aim towards no forgetting of previous knowledge, when retraining a model with additional partially similar data we should aim for performance improvement. 
This can be observed through \emph{forward knowledge transfer} -- higher performance on a new task, thanks to already incorporated knowledge, and  \emph{backward knowledge transfer} -- better generations from previous tasks when retrained on additional similar examples~\citep{lopez2017gradient}. 

To achieve this, in~\cite{deja2022multiband}, we proposed the Multiband VAE -- a method for continual learning of Variational Autoencoder that allows for continual knowledge accumulation in the autoencoder's latent space. In this work, we postulate that this approach introduced for VAE can be extended to the general framework that we call \ours{} and overview in Fig.~\ref{fig:teaser}.
The core idea behind our method is to use a generative model to continually consolidate knowledge from the stream of data. We propose to split the process of model continual training into two steps:
(1) a local encoding of data from the new task into a new model's latent space and (2) a global rearrangement and consolidation of new and previous data. 
In particular, we propose to align local data representations from consecutive tasks through the additional neural network. 

Our initial results with Variational Autoencoders indicated that this approach effectively prevents forgetting. At the same time, the model retains its plasticity and improves when updated with a similar distribution. 
Nevertheless, VAEs are known to have limited performance when used with more complex tasks. Therefore, in this work, we first show the generality of our approach by extending it to Generative Adversarial Networks. Secondly, we evaluate how we can benefit from the aligned latent representations outside the generative modeling task -- e.g., in classification. To that end, we split the classifier into two parts -- a feature extractor and a classifier head. We align representations learned from a feature extractor with the aligned representations of our generative model and train a simple classifier head from aligned representations. Experiments show that such an approach leads to better results than recent state-of-the-art generative replay methods. The contributions of this work can be summarised as follows:
\begin{itemize}
    \item We introduce the \ours{} -- a new framework for continual learning based on the alignment of representations in generative models
    \item We evaluate our approach on generative modeling task with Variational Autoencoders and Generative Adversarial Networks
    \item We show how we can benefit from generative-model-based continually aligned knowledge in the classification task 
\end{itemize}

\section{Preliminaries}

\subsection{Variational Autoencoder}
\label{sec:vae}

In this work, we follow the standard VAE training approach, introduced by~\cite{kingma2014autoencoding} that maximizes the variational lower bound of log-likelihood:
\begin{equation}
    \max_{\theta,\phi} \mathbb{E}_{q(\mathbf{\lambda} | \mathbf{x})}[\log p_\theta(\mathbf{x} | \mathbf{\lambda})]-D_{K L}(q_\phi(\mathbf{\lambda} |\mathbf{x}) \|  \mathcal{N}(\vec{0}, I))).
\end{equation}
where $\theta$ and $\phi$ are weights of the encoder and decoder, respectively.

\subsection{Wasserstein Generative Adversarial Networks}
\label{sec:gan}
Analogously, we use the Wasserstein GAN ~\cite{arjovsky2017wasserstein} with gradient penalty \cite{gulrajani2017improved}, which improves the stability of standard GAN training by optimizing the following objective. Let $\mathbf{x}$ denote the data samples and $\Tilde{\mathbf{x}}$ denote the generations of a model.
Then, the loss function of the WGAN can be defined as:
\begin{equation}
\label{eq:gan}
\min_{G_{\theta}} \max_{D_{\phi}\in{\mathcal{D}}} \mathop{\mathbb{E}}_{\mathbf{x}\sim{\mathbb{P}_{r}}}[D_{\phi}(\mathbf{x})] - \mathop{\mathbb{E}}_{\Tilde{\mathbf{x}}\sim{\mathbb{P}_{G_{\theta}}}}[D_{\phi}(\Tilde{\mathbf{x}})],
\end{equation}

\noindent where $\mathcal{D}$ is the set of 1-Lipschnitz functions, $G_{\theta}$: generator model parametrized with weights vector $\theta$, $D_{\phi}$: critic model parametrized with weights vector $\phi$, $\mathbb{P}_{r^i}$ and $\mathbb{P}_{G_{\theta}}$ are data empirical distribution, and generator's distributions respectively. Under an optimal critic, by minimizing the loss function with respect to the generator's parameters, we are thereby minimizing the Wasserstein distance between distributions $\mathbb{P}_{r}$ and $\mathbb{P}_{G_{\theta}}$.

Following ~\cite{gulrajani2017improved} we define the critic's loss function $L_{D_{\phi}}$ as:
\begin{equation}
    L_{D_{\phi}} = \mathop{\mathbb{E}}_{\Tilde{\mathbf{x}}\sim{\mathbb{P}_{G_{\theta}}}}[D_{\phi}(\Tilde{\mathbf{x}})] - \mathop{\mathbb{E}}_{\mathbf{x}\sim{\mathbb{P}_{r}}}[D_{\phi}(\mathbf{x})] + 
    \lambda \mathop{\mathbb{E}}_{\mathbf{\hat{x}}\sim{\mathbb{P}_{\mathbf{\hat{x}}}}}
    [(|| \nabla_{\mathbf{\hat{x}}}D_{\phi}(\mathbf{\hat{x}}) ||_2 - 1)^2],
\end{equation}
where $\mathbb{P}_{\mathbf{\hat{x}}}$ is the sampling distribution defined by uniform sampling along straight lines between pairs of points sampled from $\mathbb{P}_{r}$ and $\mathbb{P}_{G_{\theta}}$.

From equation~\ref{eq:gan}, we can also extract the generator's loss function $L_{G_{\theta}}$ as:
\begin{equation}
    L_{G_{\theta}} = -\mathop{\mathbb{E}}_{\Tilde{\mathbf{x}}\sim{\mathbb{P}_{G_{\theta}}}}[D_{\phi}(\Tilde{\mathbf{x}})],
\end{equation}

\section{\label{sec:related}Related Works}

The most prominent problem of continually trained systems is the abrupt performance loss when retraining the model with new data known as \emph{catastrophic forgetting}. However, in this work, we follow recent works~\cite{hadsell2020embracing,pascanu2021} that highlight the need for maximal forward and backward transfer, which means that learning from one task should facilitate learning from similar past and future tasks. 
The scope of Continual Learning (CL) strategies can be divided into three groups. The regularisation-based methods~\cite{kirkpatrick2017overcoming, zenke2017continual, li2017learning} aim to find the information crucial for previous tasks and update the model, trying not to overwrite this information. Dynamic architecture methods ~\citep{2018xu, 2017yoon} build different model versions for different tasks. Finally, replay-based approaches retrain the model on data from a new task while mixing it with additional samples from previous tasks usually stored in the memory buffer~\cite{prabhu2020gdumb, lopez2017gradient}. Novel methods employ generative models instead of the buffer to learn the data distribution in consecutive tasks.

\paragraph{Continual learning with generative rehearsal}

Most of the works incorporating generative models in continual learning relate to an idea known as generative rehearsal.  
In this technique, the base model is trained with a mixture of new data examples from the current task and recreation of previous samples generated by a generative model. 
In~\cite{2017shin+3}, authors introduced this concept with Generative Adversarial Networks (GAN). However, since the generative model can also suffer from catastrophic forgetting, the authors proposed training it with the self-rehearsal method called Generative Replay (GR). Similarly to~\cite{2017shin+3}, in \cite{van2018generative}, authors introduce a generative rehearsal model based on the Variational Autoencoder (VAE). Additionally, they combine the generative model with the base classifier to reduce the cost of model retraining. A similar approach was introduced in \cite{scardapane2020pseudo} with normalizing flows, in \cite{rostami2019complementary} with the Gaussian Mixture Model, and in DDGR~\cite{geo2023ddgr} with diffusion-based generative models. In \cite{lesort2019generative}, Lesort et al. overview different generative models trained with the GR method.  

Apart from methods that aim to generate original input from the previous tasks, a branch of works focus on rehearsing the internal data representations instead. This idea was introduced in Brain Inspired Replay (BIR)~\cite{van2020brain}, where authors propose to replay data latent representations obtained from the first few layers of a classifier, commonly known as \emph{feature extractor}. For simple datasets, authors propose to train the feature extractor alongside the classifier, while for more complex ones, they pre-train and freeze the first few convolutional layers. The idea of feature replay is further explored by~\cite{kemker2017fearnet}, where feature replay is divided into short and long-term parts. In~\cite{liu2020generative}, authors introduce GFR, which combines feature replay with feature distillation. Our extension of \ours{} to the classification task is similar to the feature replay idea. However, we do not freeze the feature extractor but train it to align with a latent space of the continually trained generative model. 

\paragraph{Continual learning of generative models}

While previous methods primarily relate to the idea of using generative models as a source of rehearsal examples for the classifier, there is a branch of methods that focuses on the problem of continual generative modeling. In  \cite{nguyen2017variational}, authors adapt regularization-based methods such as Elastic Weight Consolidation  (EWC)~\citep{kirkpatrick2017overcoming}, and Synaptic Intelligence (SI)~\citep{zenke2017continual} to the continual learning in generative models regularizing the adjustment of the most significant weights. The authors also introduce Variational Continual Learning (VCL), adjusting parts of the model architecture for each task.

In HyperCL~\cite{von2019continual} an entirely different approach is proposed, where a hypernetwork generates the weights of the continually trained model. This yields state-of-the-art results in discriminative models task-incremental training but is also applicable to the generative models. 
In order to differentiate tasks, in CURL~\cite{rao2019continual}, authors propose to learn task-specific representation and deal with task ambiguity by performing task inference within the generative model. This approach directly addresses the problem of forgetting by maintaining a buffer for original instances of poorly approximated samples and expanding the model with a new component whenever the buffer is filled. 
In BooVae~\cite{egorov2021boovae} a new approach for continual learning of VAE is proposed with an additive aggregated posterior expansion. 
Several works train GANs in the continual learning scenarios either with memory replay ~\citep{wu2018memory}, with the extension to VAEGAN in Lifelong-VAEGAN~\cite{ye2020learning}. In~\cite{zajkac2023exploring}, we studied continual learning of diffusion-based generative models ~\citep{sohl2015deep}.

\paragraph{Continual learning with disentanglement}
In VASE~\cite{achille2018life}, authors propose a method for continual learning of shared disentangled data representation.
While encoding images with a standard VAE, VASE also seeks shared generative factors. A similar concept of mixed-type latent space was introduced in LifelongVAE~\citep{ramapuram2020lifelongvae}, where it is composed of discrete and continuous values. In this work, we also use a disentanglement method with binary latent space.

\section{Method}

In this section, we introduce ~\ours{} -- a method for consolidating knowledge in a continually trained generative model. We propose to split generative replay training into two parts: (1) local training that allows us to build new data representations of the generative model and (2) global training where we align a newly trained representation to the already trained global model. Moreover, as a part of the global training, we propose a controlled forgetting mechanism where we replace selected reconstructions from previous tasks with currently available data. For simplicity, we first introduce the method for Variational Autoencoders that we call Multiband VAE, while in Sec.~\ref{sec:multiband_gan}, we propose the adaptation of our method to Generative Adversarial Networks.

\subsection{Knowledge Acquisition -- Local Training}

In the local training, we learn a new data representation band by training a VAE using only currently available data. 
Let $x_{j}^{i}$ denote the $j$-th sample of $i$-th task.
Then, for given sample $x_{j}^{i}$, and latent variable $\lambda_{j}^{i}$ we use a decoder $p_\Eparam$, which is trained to maximize posterior probability $p(x_{j}^{i}|\lambda_{j}^{i})$.  
To get the latent variable $\lambda_{j}^{i}$, we use encoder $q_\phi$ parametrized with weights $\phi$ that approximates probability $q(\lambda_{j}^{i}|x_{j}^{i})$. 
We train the VAE by maximizing ELBO as described in Sec~\ref{sec:vae}.
In the first task, this is the only part of the training, after which the local decoder becomes a global one. In other cases, we drop the local decoder after consolidation. 
\subsection{Shared Knowledge Consolidation}

\begin{figure}[!htbp]
\centering
\includegraphics[width=0.7\linewidth]{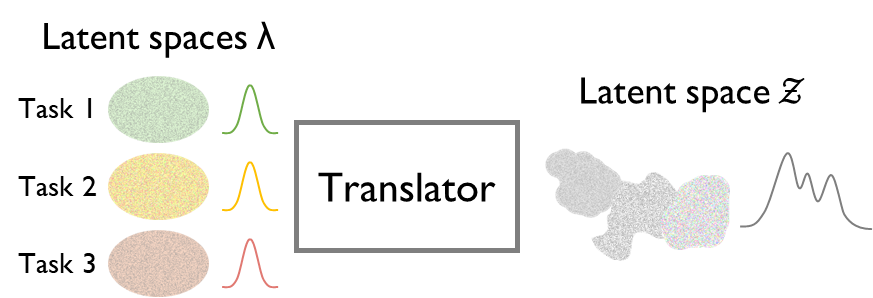}
\caption{Our translator maps individual regularized latent spaces $\lambda$ created by different local models to one global latent space $\mathcal{Z}$, where examples are stored independently of their source task.}
\label{fig:band_alignment} 
\end{figure}

In the second -- global part of the training, we align the newly trained band with already encoded knowledge.
The simplest method to circumvent interference between bands is to partition the latent space of VAE and place new data representation in a separate area of latent space (e.g., orthogonally to the previous ones). 
However, such an approach limits information sharing across separate tasks and hinders forward and backward knowledge transfer. 
Therefore, in \ours{}, we propose to align local latent representations through an additional neural network that we call \emph{translator}. The translator maps latent representations from task $i$ with their task identity 
into the common -- global latent space, where examples are stored independently of their source task, as presented in Fig~\ref{fig:band_alignment}. 

To that end, we define a translator network $t_{\rho}(\lambda^{i}, i)$ that learns a common alignment of separate latent spaces $\lambda^{i}$ conditioned with task id $i$ to a single latent variable $\mathcal{Z}$, where all examples are represented independently of their source task. Finally, we propose a global decoder $p_\Dparam(\mathbf{x}|\mathcal{Z})$
that is trained to reconstruct the original data examples $\mathbf{x}$ from latent variables $\mathcal{Z}$. 

\begin{figure}[!htbp]
\centering
\includegraphics[width=0.75\linewidth]{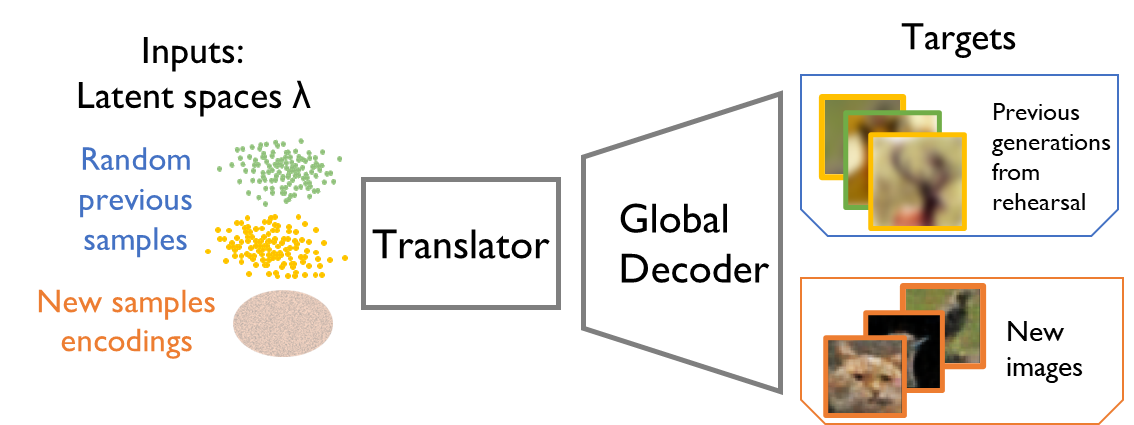}
\caption{We train our translator and global decoder with new data encoded to latent space $\lambda$ associated with original images and samples of previous data generations generated in a standard rehearsal schema.} 
\label{fig:knowledge} 
\end{figure}

To prevent forgetting, when training the translator and global decoder, we use auto-rehearsal as in standard generative replay, with a copy of the translator and decoder frozen at the beginning of the task as presented in Fig.~\ref{fig:knowledge}. As training pairs, we use a combination of original images $\mathbf{x}$ with their encodings from local encoder $\lambda$, and for previous tasks, random noise $\xi$ with generations $\Tilde{\mathbf{x}}$ obtained with a frozen translator and global decoder.  

We start translator training with a frozen global decoder to find the best fitting part of latent space $\mathcal{Z}$ for a new band of data without disturbing previous generations. To that end, we minimize the reconstruction loss: 
\begin{equation}
    \min_\rho \Big[ \sum_{i=1}^{k-1} \underbrace{||\Tilde{\mathbf{x}}^i- p_\omega(t_\rho(\xi,i))||^2_2}_\text{previous task $i$}  + \underbrace{||\mathbf{x}^k- p_\omega(t_\rho(\lambda,k))||^2_2}_\text{current task} \Big], 
\end{equation} 

where $k$ is the number of all tasks.

Then, we optimize the translator and global decoder jointly, minimizing the reconstruction error between outputs from the global decoder and training examples.

\begin{equation}
    \min_{\rho,\omega} \Big[ \sum_{i=1}^{k-1} \underbrace{||\Tilde{\mathbf{x}}^i- p_\omega(t_\rho(\xi,i))||^2_2}_\text{previous task $i$} + \underbrace{||\mathbf{x}^k- p_\omega(t_\rho(\lambda,k))||^2_2}_\text{current task} \Big], 
\end{equation}

To generate new example $\Tilde{x}$ with Multiband VAE, we randomly sample task id $i \sim \mathcal{U}(\{1,\dots,k\})$ and latent representation $\lambda_{t} \sim \mathcal{N}(\vec{0}, I)$, where $k$ is the number of all tasks. These values are mapped with the translator network to latent variable $z$, which is the input to the global decoder. Therefore, the translator and global decoder are the only models that are stored in-between tasks.

\subsection{Controlled Forgetting}

\begin{figure}[!htbp]
\centering
\includegraphics[width=0.75\linewidth]{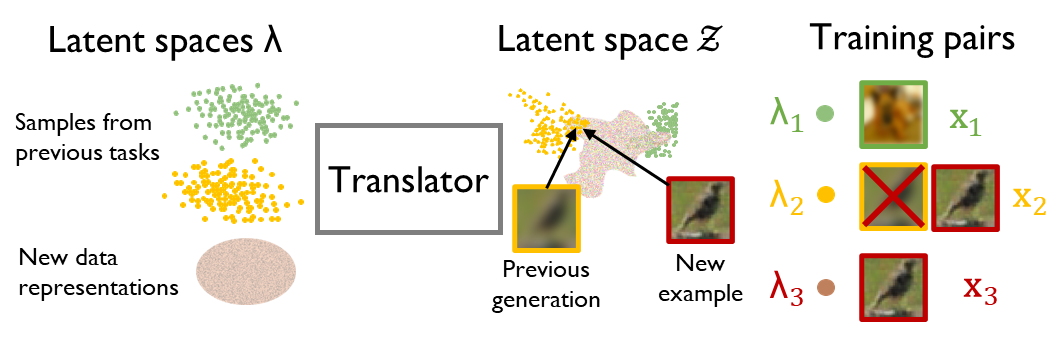}
\caption{When creating rehearsal training pairs with generations from previous data examples, we calculate the similarity between the sampled example and the closest currently available data sample in the common latent space $\mathcal{Z}$. If this similarity is above a given threshold, we allow forgetting of the previous reconstruction by substituting the target generation with a currently available similar image.} 
\label{fig:forgetting} 
\end{figure}

In a real-life scenario, it is common to encounter similar data examples in many tasks. In such a case, we would like our continuously trained model to refresh the memory of examples instead of focusing on old rehearsed memories that are likely distorted~\cite{alemohammad2023self}.
Therefore, we propose a mechanism for controlled forgetting of past reconstructions during the translator and global decoder joint training. To that end, when creating new training pairs, we compare representations of previous data generations used as new targets with representations of data samples from the current task in the common latent space $\mathcal{Z}$. If these representations are similar enough, we substitute previous data reconstruction
with the current data sample as presented in Fig.~\ref{fig:forgetting}.

More specifically, when training on task $i$, after training the translator with a frozen global decoder, we first create a subset $\mathcal{Z}^i = t_\rho(q_\phi(\mathbf{x}^i), i)$ with representations of all currently available data in the joint latent space $\mathcal{Z}$.
Now, for each data sample $x^l_j$ generated as a rehearsal target from previous task $l<i$ and random variable $\lambda_j^l$, we compare its latent representation $z_j = t_\rho(\lambda_j^l,j)$ with all elements of set $\mathcal{Z}^i$ 
\begin{equation}
    sim(z_{j}) \coloneqq \max_{z_{q} \in \mathcal{Z}^i} cos(z_{j}, z_{q}).
\end{equation}
If $sim(z_{j}) \geq \gamma$ we substitute target sampled reconstruction $x^l_j$ with the respective original image from $\mathbf{x}^i$.
Intuitively, $\gamma$ controls how much we want to forget from task to task, with $\gamma=0.9$ being a default value for which we observe a stable performance across all benchmarks. 

\subsection{Multiband training of Generative Adversarial Networks} \label{sec:multiband_gan}
The core idea behind the training of Multiband GAN is very similar to the Multiband VAE training. We also split the process of training into local and global parts. In the following sections, we describe the key differences between the GAN and VAE Multiband training.

Analogously to VAE's case, in local training of Multiband GAN for each task $i$, we follow standard WGAN training procedure, described in Sec.~\ref{sec:gan}, using only currently available data samples $\mathbf{x}^i$.  
In the first task, this is the only part of the training, after which the local generator becomes a global one. In every other task, we start training both local models from weights pretrained on the previous task.

\begin{figure}[!htbp]
    \centering \includegraphics[width=0.8\linewidth]{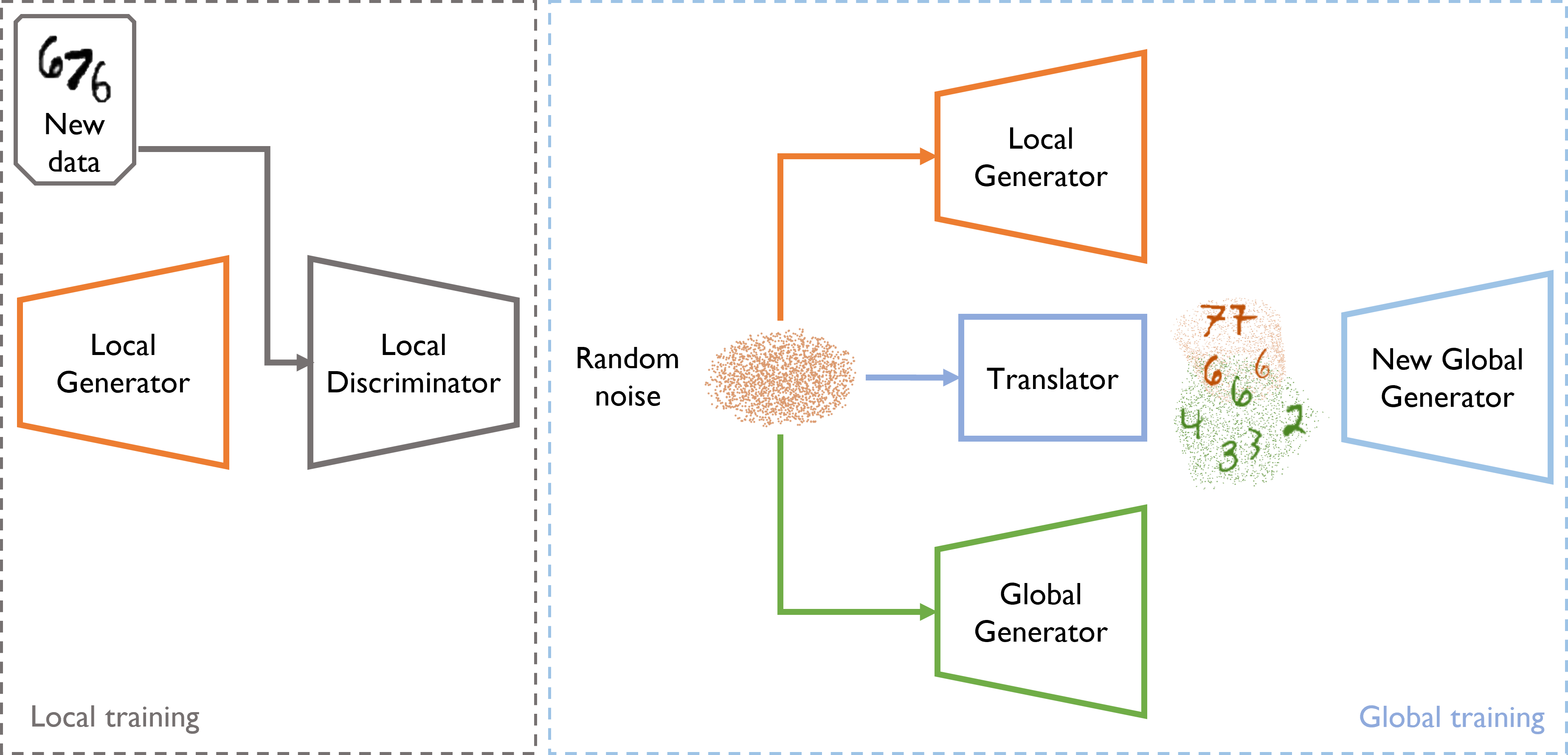}
    \caption{Overview of the Multiband GAN. With each new task, we first learn a \textit{local} copy of our model using only new data examples. Then, we consolidate it with our current global generator - the main model that is able to generate examples from all tasks.}
    \label{fig:multiband_gan}
\end{figure}

During the global training, we distill the knowledge from the local model to the global one, which is trained on previous tasks. To that end, we define translator network $t_{\rho}(\xi, i)$ that learns a common alignment of noises $\xi$ to the global embeddings $\mathcal{Z}$. As training pairs, we use generations $\Tilde{\mathbf{x}}$ with the random noises $\xi$ from which they were sampled.

We start translator training with a frozen global generator. To that end, we minimize the loss:
\begin{equation}
    \min_{\rho} \sum_{i=1}^{k} ||\Tilde{\mathbf{x}}^i- G_\theta(t_\rho(\xi,i))||^2_2, 
\end{equation}
where $k$ is the number of all tasks and $\Tilde{\mathbf{x}}^i$ denote the generated samples from $i$-th task.

Then, we optimize the parameters of the translator and global generator jointly:
\begin{equation}
    \min_{\rho, \theta} \sum_{i=1}^{k} ||\Tilde{\mathbf{x}}^i- G_\theta(t_\rho(\xi,i))||^2_2, 
\end{equation}

\subsection{\ours{} for classification}

\begin{figure}[!htbp]
    \centering
    \includegraphics[width=.9\textwidth]{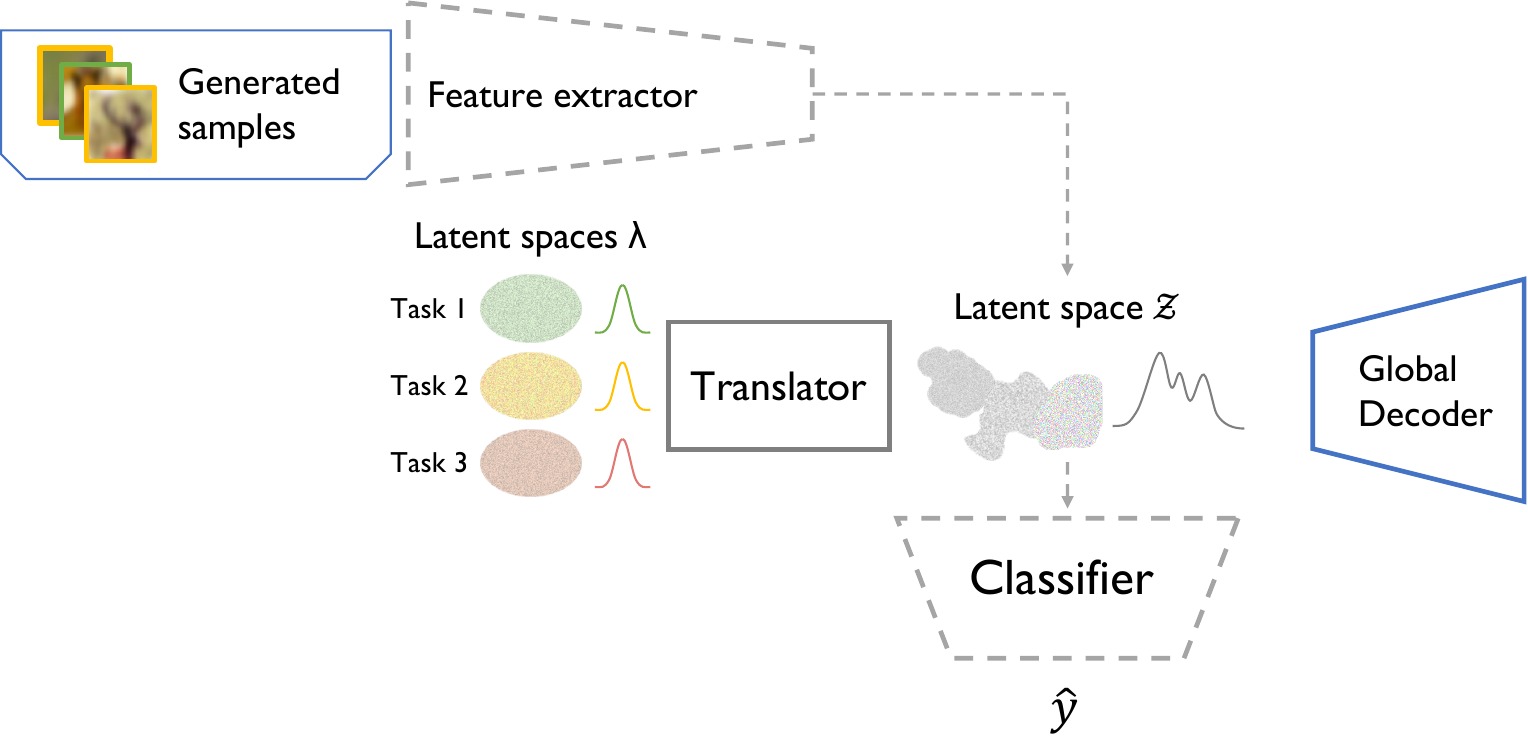}
    \caption{Application of \ours{} to downstream classification task. We train a feature extractor to encode images into latent space $\mathcal{Z}$, and a classifier head that predicts class $\hat{y}$ from latent representations $\mathcal{Z}$.}
    \label{fig:classifier_training}
\end{figure}

On top of our approach to continual learning of generative models, we propose to use a generative model as a focal point of the continually trained system, benefiting from the aligned knowledge for a different downstream task. As a concrete instantiation of this idea, we evaluate how we can use continually aligned representations for classification. In particular, we split the classifier into two parts, as visualized in Fig.~\ref{fig:classifier_training}. 
The first one, dubbed \emph{feature extractor}, is trained to encode data examples directly into the aligned latent space $\mathcal{Z}$ of the continually trained generative model. The second one is a simple classifier that assigns latent representations to classes in a standard class-incremental setup. In this section, we describe the proposed approach in detail.

\subsubsection{Feature extractor training}

We train the feature extractor to reverse the reconstruction process of the new global decoder. To that end, in each task $i$, we create a set of training pairs in the form of generations $\Tilde{\mathbf{x}}$ from each task encountered so far, with their latent representations $z$.
The feature extractor network $f_{\tau}$ with parameters $\tau$ is trained to map $\Tilde{\mathbf{x}}$ into aligned representations $z$ with MSE loss:

\begin{equation}
    L_{f_{\tau}} = {(z - f_\tau(\Tilde{\mathbf{x}}))^2}
\end{equation}

\subsubsection{Classifier training}

On top of the feature extractor, we train a simple classifier $c$ with parameters $\kappa$ that predicts the classes from aligned latent representations $\hat{y} = c_\kappa(z)$. Similarly to the feature extractor training, we train the classifier on latent representations $z$ of generations from both previous and current tasks.
As a loss function, we use cross-entropy loss:
\begin{equation}
L_c =  -\sum_{i=1}^{C} y_i \cdot \log(\hat{y}_i),
\end{equation}
\noindent where C is the number of classes

\section{\label{sec:experiments}Experiments}

In this section, we present the experimental results, comparing our continual generative modeling approach with alternative methods. Expanding on our prior work, we include additional experiments involving Generative Adversarial Networks that outperform similar approaches, including our Multiband VAE. Additionally, we show how we can use aligned latent representations from our continually trained generative models for the classification task. Comparative analyses highlight the effectiveness of our approach against state-of-the-art generative replay methods.

\subsection{Toy example}
\begin{figure*}[!htbp]
\centering
\includegraphics[width=\linewidth]{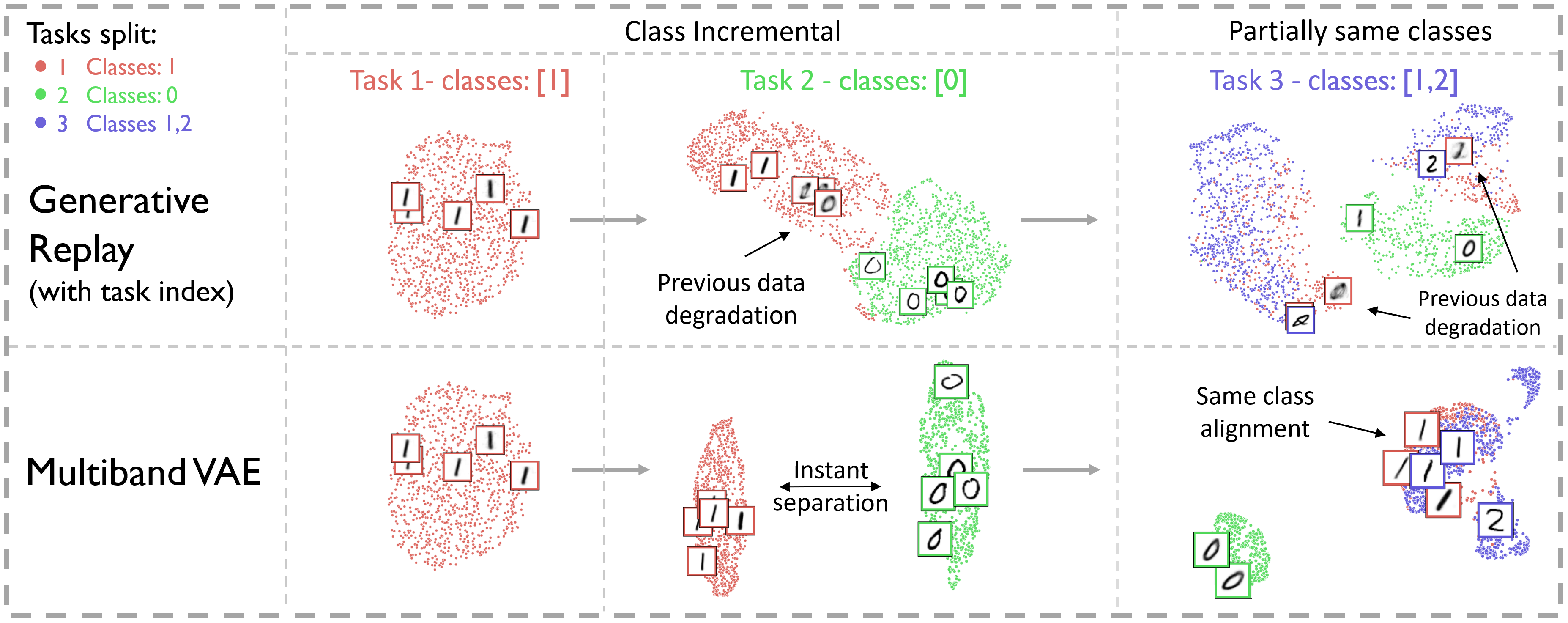}
\caption{Visualization of latent space $\mathcal{Z}$ and generations from VAE in standard Generative Replay and our multiband training for the three tasks (different colors) in a case of entirely different new data distribution, and partially same classes. GR does not instantly separate data from different tasks, which results in the deformation of previously encoded examples. On the contrary, our \ours{} can separate representations from different classes while properly aligning examples from the same new class if present.} 
\label{fig:toy_example} 
\end{figure*} 
To visualize the difference between standard Generative Replay and \ours{}, in Fig.~\ref{fig:toy_example}, we visualize the results from an experiment on a simplified setup, using only 3 tasks created from the MNIST dataset. We limit the data to only 3 classes and split them into 3 tasks:
\begin{itemize}
    \item Task 1 containing only examples from class 1
    \item Task 2 containing only examples from class 0
    \item Task 3 containing examples from class 2 and additional examples from class 1 not used in the first task
\end{itemize}
With this simplified setup, we run the whole training through the three tasks using either standard generative replay or our \ours{} method. 

When presented with data from a new distribution (different class in task 2), our method places a new band of data in a separate part of a common latent space $\mathcal{Z}$. On the contrary, the generative replay method does not separate representations of data from different tasks, what results in the deformation of previously encoded examples.
At the same time, when presented with data with partially the same classes (task 3), our translator is able to properly align bands of data representations so that similar data examples (in this case, ones) are located in the same area of latent space $\mathcal{Z}$ independently of the source task, without interfering with zeros and twos.

\subsection{Evaluation Setup}
For a fair comparison, in all evaluated methods, we use a Variational Autoencoder architecture similar to the one introduced by~\cite{nguyen2017variational}, with nine dense layers. However, our~\ours{} is not restricted to any particular architecture, so we also include experiments with a convolutional version. 

In our GAN approach, we adapt the WGAN convolutional architecture to be as similar as possible to the convolutional VAE that we use so that the number of parameters of these two models is nearly the same. 
The exact architectures and training hyperparameters for both VAE \footnote{\url{https://github.com/KamilDeja/multiband_vae}} and GAN \footnote{\url{https://github.com/cywinski/multiband_gan}} approaches are enlisted in the appendix~\ref{app:model_arch} and linked code repositories.

We do not condition our generative models with class identities since it greatly simplifies the problem of knowledge consolidation and applies to all evaluated methods. However, in the VAE approach, we use additional binary latent space trained with Gumbel softmax~\citep{jang2016categorical}, similarly to~\cite{ramapuram2020lifelongvae}.  

\begin{figure}[!htbp]
\centering
\footnotesize
\stackunder[5pt]{\includegraphics[width=0.25\linewidth]{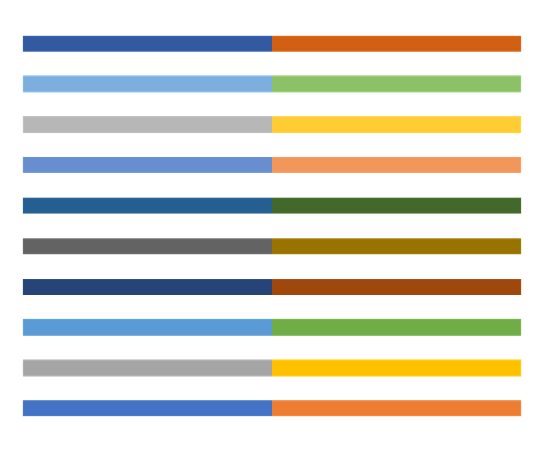}}{Class incremental}%
\hspace{1pt}
\stackunder[5pt]{\includegraphics[width=0.25\linewidth]{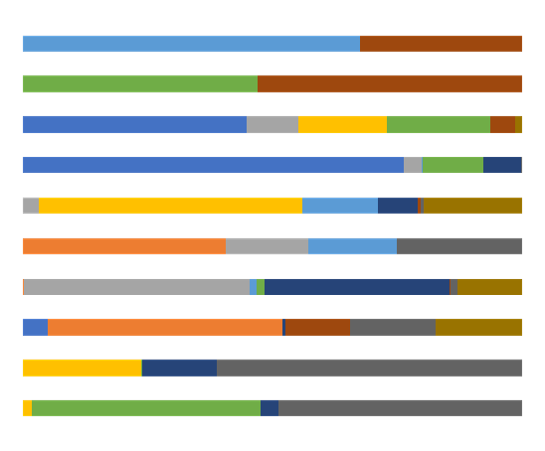}}{Dirichlet $\alpha$=$1$}%
\hspace{1pt}
\stackunder[5pt]{\includegraphics[width=0.25\linewidth]{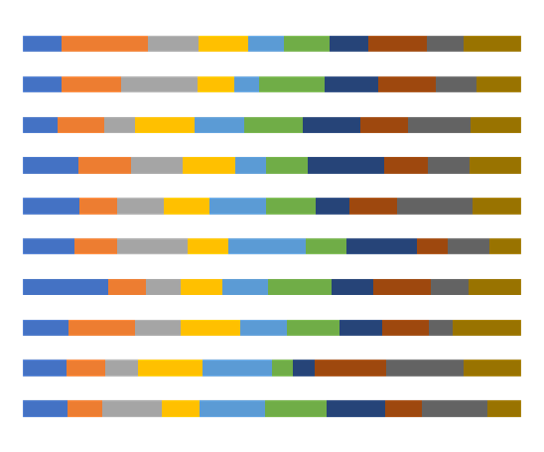}}{Dirichlet $\alpha$=$100$}%

\caption{\small \label{fig:splits} Class splits for different continual learning scenarios. In class incremental split each task consists of separate classes. For $\alpha=1$ Dirichlet distribution, we have highly imbalanced splits with randomly occurring dominance of one or two classes. For higher values of parameter $\alpha$, classes are split almost equally.}
\end{figure}

\begin{table*}[!htbp]
  \centering
  \resizebox{\textwidth}{!}{
  \begin{tabular}{l||ccc|ccc|ccc|ccc|c}
    \toprule
     & \multicolumn{3}{c}{Split-MNIST}  & \multicolumn{3}{c}{MNIST}& \multicolumn{3}{c}{Split-Fashion MNIST} & \multicolumn{3}{c}{Fashion MNIST} & CERN \\
    & \multicolumn{3}{c}{Class Incremental}&\multicolumn{3}{c}{Dirichlet $\alpha=1$ }&\multicolumn{3}{c}{Class Incremental} & \multicolumn{3}{c}{Dirichlet $\alpha=1$} & Class Inc.\\ 
    Num. tasks &\multicolumn{3}{c}{5}&\multicolumn{3}{c}{10}&\multicolumn{3}{c}{5}&\multicolumn{3}{c}{10}&5\\
    \midrule
    Measure &FID $\downarrow$&Prec $\uparrow$&Rec $\uparrow$ &FID $\downarrow$&Prec $\uparrow$&Rec $\uparrow$ &FID $\downarrow$&Prec $\uparrow$&Rec $\uparrow$ &FID $\downarrow$&Prec $\uparrow$&Rec $\uparrow$ & Wass $\downarrow$\\

    \midrule
    SI & 129 & 77& 80  &153&75&76&134&28&24&140&21&19 &21.1\\
    EWC & 136 & 73& 82 & 120&79&83&126&25&25 &137&24&22& 29.7\\
    Generative replay &120&79&87&254&70&65&96&43&58&133&35&43 & 11.1\\
    VCL & 68 & 85& 94&  127&78& 80&104&30&32 &138&21&20&24.3\\
    HyperCL & 62&91&87 & 148&78&75&108&46&33&155&35&21 &7.8\\
    CURL & 107& 95 &77 & 181&84&74 & 86&47&64 & 83&46&56 & 16.8 \\
    Livelong-VAE & 173 & 75 & 72 &224&63&73 &131&33&62 &201&9&49&7.7\\
    Livelong-VAEGAN& 48 & 98 & 89 &131 &90 &83 &78 & 54& 79 &108 &54 &64 &15.1\\
    \textbf{Multiband VAE}& 24&94&97& 41&92&96&61&66&69 &82&62&65 &\textbf{6.6}\\
    \midrule
    \textbf{Multiband VAE(conv)}&23& 92&98& 30&92&97 &56&65&72& 77&58&69&8.1\\
    \textbf{Multiband GAN (conv)}& \textbf{9}&\textbf{99}&\textbf{99}& \textbf{20}&\textbf{98}&\textbf{98} &\textbf{34}&\textbf{92}&\textbf{92} &\textbf{58}&\textbf{88}&\textbf{83}& 10.3\\
    \bottomrule
  \end{tabular}
  }
    \caption{\label{tab:results_mnist} Average FID and distribution Precision (Prec) and Recall (Rec) or Wasserstein distance between original and generated simulation channels after the final task in different data incremental scenarios. Our method with vanilla architecture outperforms competing solutions. Furthermore, we are able to significantly improve the results with convolutional architectures.
}
\end{table*}

\subsection{Evaluation}
To assess the quality of our method, we conduct a series of experiments on a broad set of common continual learning benchmarks. We prepare a set of training scenarios for each of them to evaluate various aspects of generative continual learning. This is the only time we use data classes since our solution is fully unsupervised.
We evaluate our method on MNIST, Omniglot~\citep{lake2015human}, FashionMNIST~\citep{xiao2017fashionmnist} and CelebA~\citep{liu2015faceattributes} datasets. We further conduct the experiments on CIFAR10 and CIFAR100 \citep{krizhevsky2009learning} datasets only for our GAN-based approach since the performance of VAE-based architectures is very limited on such complex datasets.

\begin{table*}[tb]
  \centering
     \resizebox{\textwidth}{!}{
      \begin{tabular}{l||ccc|ccc|ccc|ccc|ccc}
        \toprule
         & \multicolumn{3}{c}{Split-Omniglot}  & \multicolumn{3}{c}{Split-Omniglot}& \multicolumn{3}{c}{Omniglot} & \multicolumn{3}{c}{FashionM$\rightarrow$MNIST}&\multicolumn{3}{c}{MNIST$\rightarrow$FashionM} \\
        & \multicolumn{3}{c}{Class Incremental}&\multicolumn{3}{c}{Class Incremental}&\multicolumn{3}{c}{Dirichlet $\alpha=1$} & \multicolumn{3}{c}{Class Incremental}&\multicolumn{3}{c}{Class Incremental}\\ 
        Num. tasks &\multicolumn{3}{c}{5}&\multicolumn{3}{c}{20}&\multicolumn{3}{c}{20}&\multicolumn{3}{c}{10}&\multicolumn{3}{c}{10}\\
        \midrule
        Measure &FID$\downarrow$&Prec$\uparrow$&Rec$\uparrow$&FID$\downarrow$&Prec$\uparrow$&Rec$\uparrow$&FID$\downarrow$&Prec$\uparrow$&Rec$\uparrow$&FID$\downarrow$&Prec$\uparrow$&Rec$\uparrow$&FID$\downarrow$&Prec$\uparrow$&Rec$\uparrow$\\
        \midrule
        SI&48&87&81&115&64&28&140&18&16&146&18&15&157&21&19\\
        EWC& 46&88&81&106&68&31 &106&74&38&119&72&30&133&25&23\\
        Generative replay&45&88&82&74&72&62&92&75&53&99&36&45&111&24&39\\
        VCL&48&87&82& 122&62&21&127&71&25&81&45&51&79&45&55\\
        HyperCL& 54&86&76 &98&86&45&115&84&38&128&31&28&143&30&28 \\
        CURL& 22&95&95 &31&\textbf{96}&92& 26&94&92 & 98&69&42 & 122&47&37\\
        Lifelong-VAE& 49&87&83&  79&83&59&  93&83&51& 173&13&50& 200&12&52\\
        Lifelong-VAEGAN& 31&96&90& 71&83&70& 63&85&78& 127&34&61& 91 &52&73 \\
        \textbf{Multiband VAE}& 21 & 97 & 93&33& 95&86&41&95&83 &51&65 &70& 49&67&73 \\
        \midrule
        \textbf{Multiband VAE (conv)}&12&98&96&24&95&91&24&\textbf{96}&91&49&68&70&49&70&70\\
        \textbf{Multiband GAN (conv)}& \textbf{1} & \textbf{99} & \textbf{98} & \textbf{3} & \textbf{96} & \textbf{95} &\textbf{4}&\textbf{96}&\textbf{93} &\textbf{31}&\textbf{89} &\textbf{87}& \textbf{26}&\textbf{93}&\textbf{91} \\
        \bottomrule
  \end{tabular}
}
    \caption{  \label{tab:results_omni} Average Fréchet Inception Distance (FID) and distribution Precision (Prec) and Recall (Rec) after the final task in different data incremental scenarios. 
    In more challenging datasets \ours{} outperforms competing solutions.
    }
\end{table*}

To assess whether models trained according to our method suffer from catastrophic forgetting, we run class incremental scenarios introduced by~\cite{van2019three}. Although CI is a very difficult continual learning scenario for discriminative models, it actually simplifies the problem of learning data distribution in the generative model's latent space because the identity of the task conditions final generations. Thus, generative models can learn separate data representations for each task. Therefore, we also introduce more complex data splits with no assumption of independent task distributions. To that end, we split examples from the same classes into tasks according to the probability $q\sim Dir(\alpha p)$ sampled from the Dirichlet distribution, where $p$ is a prior class distribution over all classes, and $\alpha$ is a \textit{concentration} parameter that controls similarity of the tasks, as presented in Fig.~\ref{fig:splits}. 
In particular, we exploit the Dirichlet $\alpha=1$ scenario, where the model has to learn the differences between tasks while consolidating representations for already known classes. In such a scenario, we expect forward and backward knowledge transfer between tasks for the model to learn the entire distribution of the data.

We use the Fréchet Inception Distance (FID)~\citep{heusel2017gans} to measure the quality of generations from different methods. 
As proposed by~\cite{binkowski2018demystifying}, for simpler datasets, we calculate FID based on the LeNet classifier pre-trained on the whole target dataset. 
Additionally, we report the precision and recall of the distributions as proposed by~\cite{sajjadi2018assessing}. As the authors indicate, those metrics disentangle the FID score into two aspects: the quality of generated results (Precision) and their diversity (Recall).

For each experiment, we report the FID, Precision, and Recall averaged over the final scores for each task separately. For methods that do not condition generations on the task index (CuRL and LifelongVAE), we calculate measures in comparison to the whole test set. The results of our experiments are presented in Tab.~\ref{tab:results_mnist} and Tab.~\ref{tab:results_omni}, where we show scores averaged over three runs with different random seeds. 

We also use real data from detector responses in the LHC experiment to compare different continual learning generative methods in real-life scenarios. Calorimeter response simulation is one of the most profound applications of generative models where those techniques are already employed in practice ~\citep{paganini2018calogan}. In our studies, we use a dataset of real simulations from a Zero Degree Calorimeter in the ALICE experiment at CERN introduced by~\cite{deja2020end}, where a model is to learn outputs of $44 \times 44$ resolution energy depositions in a calorimeter. Following~\cite{deja2020end}, instead of using FID for evaluation, we benefit from the nature of the data and compare the distribution of real and generated channels -- the sum of selected pixels that well describe the physical properties of simulated output. We measure generations' quality by reporting the Wasserstein distance between the original and generated channel distributions. We prepare a continual learning scenario for this dataset by splitting examples according to their input energy, simulating changing conditions in the collider. In practice, such splits lead to continuous change in output shapes with partial overlapping between tasks -- similar to what we can observe with Dirichlet-based splits on standard benchmarks (see appendix~\ref{app:cern} for more details and visualizations).

As presented in Tab.~\ref{tab:results_mnist}, our standard setup that uses linear VAE architecture outperforms comparable methods regarding the quality of generated samples. Furthermore, we are able to further improve the quality of generations by introducing convolutional VAE and GAN architectures. 

\begin{figure}[tbp]
\centering
\footnotesize

\stackunder[5pt]{\includegraphics[width=0.3\linewidth]{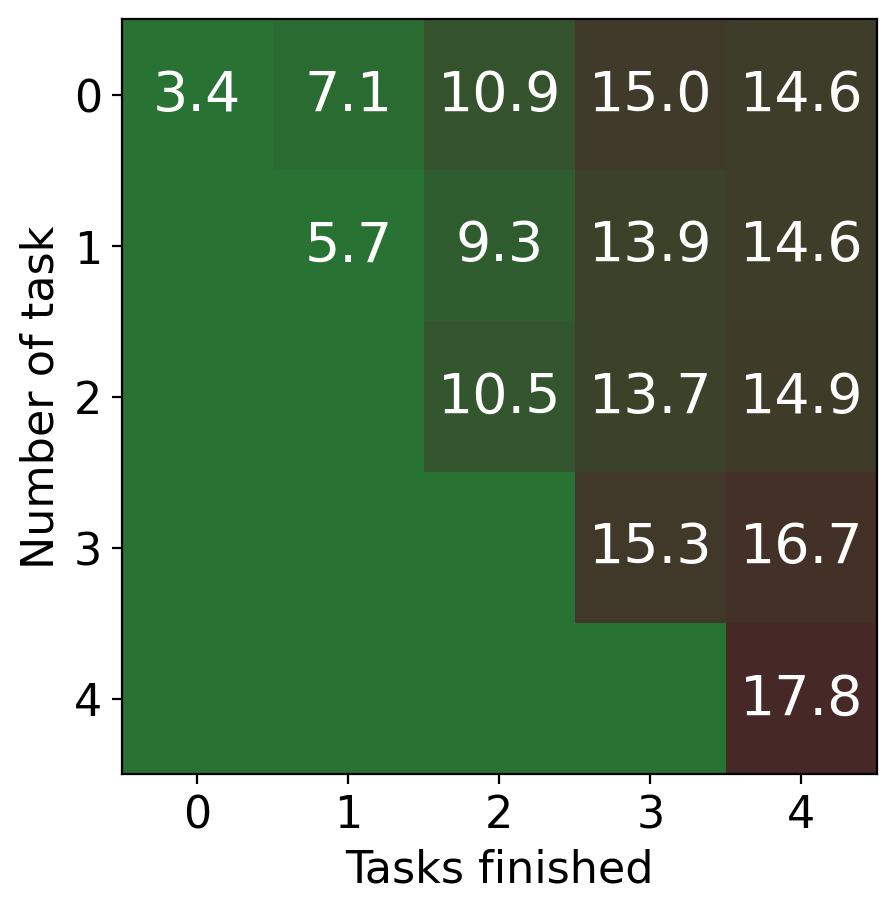}}{\centering Generative replay}%
\hspace{1pt}
\stackunder[5pt]{\includegraphics[width=0.3\linewidth]{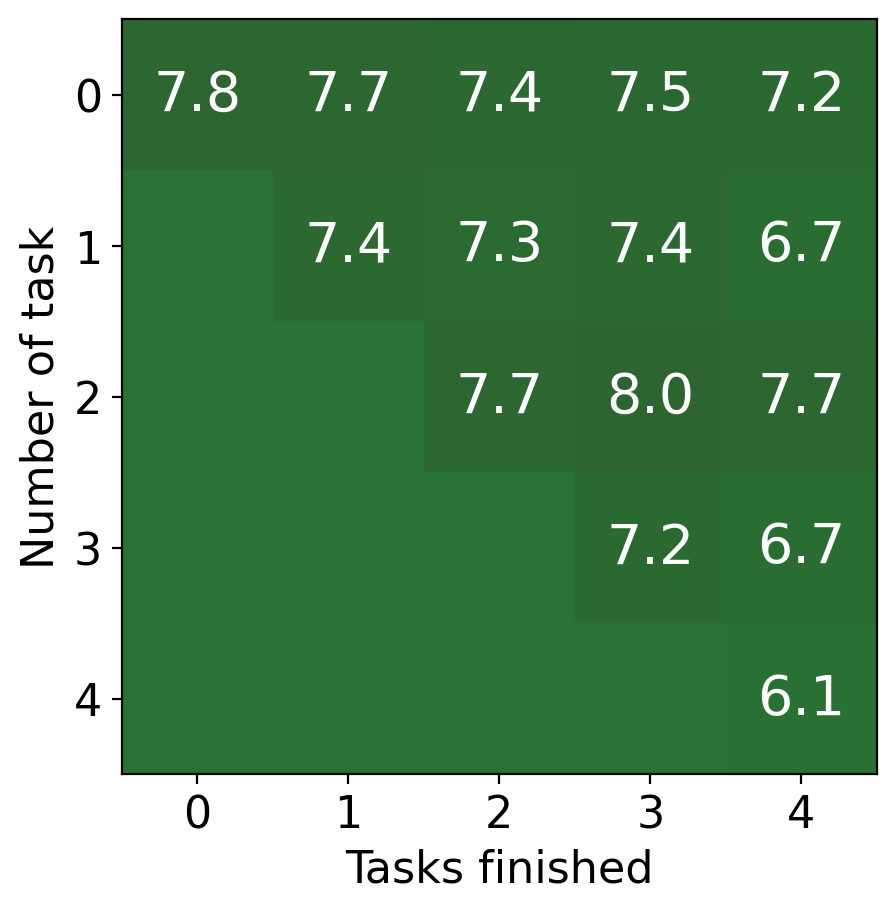}}{\centering \ours{}}%

\caption{\small \label{fig:cern_example} Comparison of Wasserstein distance $\downarrow$ between original simulation channels and generations from VAE trained with standard GR and our method. \ours{} well consolidates knowledge with visible forward transfer (each row starts with a better score) and backward transfer (improvement for some rows when retrained with more data). At the same time, standard GR struggles to retain the quality of generations on old tasks.}
\end{figure}

Results of comparison on the Omniglot dataset with 20 splits (Tab.~\ref{tab:results_omni}) indicate that for almost all of the related methods, training with the data splits according to the Dirichlet $\alpha=1$ distribution poses a greater challenge than the class incremental scenario. However, our~\ours{} can precisely consolidate knowledge from such complex setups while still preventing forgetting in CI scenarios. This is only comparable to CURL, which achieves this goal through additional model expansion. Here, we also observe the superiority of the GAN approach, which can learn the Omniglot dataset's data distribution much more precisely than any other compared method.  

Experiments on the DoubleMNIST dataset, where examples are introduced from one dataset after another, indicate the superiority of~\ours{} over similar approaches. Moreover, we can see the impact of class order on the quality of samples from the final model \citep{masana2020class}. Our approach, especially GAN-based, tends to perform better when it starts from simpler tasks (MNIST) than more difficult ones (FashionMNIST).
In the real-life CERN scenario, our model also clearly outperforms other solutions. In Fig.~\ref{fig:cern_example}, we present how generations quality for this dataset changes in standard generative replay and \ours{}, showing both forward and backward knowledge transfer in our method.

Finally, we evaluate our approach on a more complex CelebA dataset as a benchmark, with over 200 000 images of celebrity faces in 64x64 resolution. Based on annotated features, we split the dataset into 10 classes based on the hair color/cover (blonde, black, brown, hat, bald or gray). 
\begin{table*}[t!]
  \centering
  \resizebox{\textwidth}{!}{
  \begin{tabular}{l||ccc|ccc|ccc|ccc}
    \toprule
    CelebA split& \multicolumn{3}{c}{Class Incremental}&\multicolumn{3}{c}{Dirichlet $\alpha=1$}&\multicolumn{3}{c}{Dirichlet $\alpha=100$}&\multicolumn{3}{|c}{Upper bound}\\ 
    Num. tasks &\multicolumn{3}{c}{5}&\multicolumn{3}{c}{10}&\multicolumn{3}{c}{10}&\multicolumn{3}{|c}{1}\\
    \midrule
    Measure &FID$\downarrow$&Prec$\uparrow$&Rec$\uparrow$&FID$\downarrow$&Prec$\uparrow$&Rec$\uparrow$&FID$\downarrow$&Prec$\uparrow$&Rec$\uparrow$&FID$\downarrow$&Prec$\uparrow$&Rec$\uparrow$\\
    \midrule
    Separate models (VAE) &103&31&21&105&24.5&7.6&109&28.4&10.6&88&35&30\\
    Generative Replay&105&23.4&14.9&109&14.6&7.4&102&17.2&11.6&88&35&30\\
    \textbf{Multiband VAE}&95&28.5&23.2&93&33&22&89&36.2&28&88&35&30\\
    \textbf{Multiband GAN} & \textbf{39} & \textbf{81} & \textbf{79} &\textbf{60}&\textbf{67}&\textbf{67}&\textbf{37}&\textbf{80}&\textbf{83} &\textbf{19}&\textbf{91}&\textbf{93} \\
    \bottomrule
  \end{tabular}
  }
  \caption{\label{tab:results_celeba} Average FID, distribution Precision, and Recall after the final task on the CelebA dataset. Our \ours{} consolidates knowledge from separate tasks even in the class incremental scenario, clearly outperforming other solutions. With more even splits, our method converges to the upper bound, which is a model trained with full data availability.}
\end{table*}

In Tab.~\ref{tab:results_celeba}, we show the results of experiments with this dataset split in class incremental and Dirichlet scenarios. 
For class incremental scenario, \ours{} learns to separate bands of examples from different tasks with disjoint distributions, while results improve if in training scenario model is presented with more similar examples. In the latter case, with Dirichlet $\alpha=100$ splits, 
the vanilla VAE model reaches the quality of its upper bound, which is a standard Variational Autoencoder trained with full access to all examples in the stationary training. 
On this challenging benchmark, we see a significant improvement in the quality of GAN's generated samples compared to the VAE model. In each tested CelebA scenario, the GAN-based approach achieves over two times better precision and recall.

\subsection{Memory Requirements and Complexity}

The memory requirements of \ours{} are constant and equal to the size of the base models (either VAE or GAN) with an additional translator, which is a small neural network with 2 fully connected layers. When training on the new task, our method requires additional temporary memory for the local model freed when finished. This is contrary to similar methods (HyperCL, VCL, CURL) 
which have additional constant or growing memory requirements.
The computational complexity of our method is the same as that of methods based on generative rehearsal (VCL, LifelongVAE, Lifelong-VAEGAN). In experiments, we use 
\begin{wraptable}{r}{0.4\textwidth}
\vskip  -0.3cm
    \begin{adjustbox}{center}
  \centering
  \small
  \begin{tabular}{l|l}
    \toprule
    Modification & FID$\downarrow$\\
    \midrule
    Generative replay & 254 \\ 
    + Two-step training &64\\
    + Translator &53\\
    + Binary latent space &44\\
    + Controlled forgetting & 41\\
    + Convolutional model & 30\\
    \bottomrule
  \end{tabular}
\end{adjustbox}
\caption{ \label{tab:ablation_studies}
Ablation study on the MNIST dataset with Dirichlet $\alpha=1$ distribution. Average FID  after the last task.}

\vskip  -1.8cm
\end{wraptable}
\noindent the same number of epochs for all methods, while for~\ours{}, we split them between local and global training. 

\subsection{Ablation study}

\vspace{0.3cm}
The main contribution of this work is our two-step training procedure, yet we also introduce several mechanisms that improve knowledge consolidation. 
Tab.~\ref{tab:ablation_studies} shows how those components contribute to the final score in the case of Multiband VAE training.

\vspace{0.2cm}
\subsection{Using latent representations for classification}

On top of our primary evaluation of our model's generative capabilities, we introduce a proof-of-concept extension of \ours{} to the classification problem. In table \ref{tab:results}, we provide the evaluation results on different datasets, subdivided by the architecture - VAE or GAN. We also specify the experiment scenario alongside the exact network architecture. Training code and hyperparameters are enlisted in the code repository \footnote{\url{https://github.com/jrx-napoli/cl_classifier}}.

In table \ref{tab:comparison}, we compare our approach to other state-of-the-art generative-replay methods and regularisation-based continual learning classification benchmarks. As visible, our \ours{} clearly outperforms competing methods by a large margin (especially on more complex datasets such as CIFAR10 and CIFAR100). Contrary to recent methods, our model is able to continually align knowledge from separate portions of data without the need for pretraining on external data~\cite{van2020brain,geo2023ddgr} or enlarged first task~\cite{liu2020generative,geo2023ddgr}. 
Additional results in different setups, including the Dirichlet-based knowledge consolidation scenarios, are presented in Appendix~\ref{app:classification_results}. For all experiments, we use network architectures as specified in table \ref{tab:results}.

\begin{table}[H]
    \centering
    \small
\begin{tabular}{l|ccc}
\multirow{2}{*}{Method}  & MNIST & CIFAR10 & CIFAR100 \\ 
& 5 tasks & 5 tasks & 20 tasks \\ \midrule
\multicolumn{4}{c}{No Memory} \\ \midrule
Finetune & 18.8 $\pm$ 0.5 & 15.0 $\pm$ 3.1 & 5.9 $\pm$ 0.15  \\
SI \cite{zenke2017continual}& 19.7 $\pm $ 0.1  & 19.5 $\pm $ 0.2 & 13.3 $\pm $ 1.14 \\
MAS \cite{aljundi2018memory} & 19.5 $\pm $ 0.3 & 15.0 $\pm$ 2.2 & 10.4 $\pm $ 0.80\\
EWC \cite{kirkpatrick2017overcoming}& 19.8 $\pm $ 0.1  & 19.5 $\pm $ 0.1 & 9.5 $\pm $ 0.83\\

LwF \cite{li2017learning} & 24.2  $\pm $ 0.3 & 19.6 $\pm $ 0.1 & 14.1 $\pm $ 0.87\\

\midrule
\multicolumn{4}{c}{Generative Replay} \\ \midrule
    DGR~\cite{2017shin+3} & 91.2 $\pm $ 0.3 & 27.3 $\pm$ 1.3 &7.41 $\pm $ 0.13 \\
    DGR+distill~\cite{van2019three} & 91.8 $\pm $ 0.3 & 28.4 $\pm$ 0.3 & 7.28 $\pm $ 0.40\\
    RTF~\cite{van2018generative} & 92.6 $\pm $ 0.2 & 28.7 $\pm$ 0.2 & 7.88 $\pm $ 0.54\\
    BIR* \cite{van2020brain}  & 92.8 $\pm $ 1.1& 20.7 $\pm$ 0.2 & 6.38 $\pm$ 0.65 \\
    GFR \cite{liu2020generative} & - & 26.7 $\pm$ 1.9 & 9.5 $\pm$ 0.59  \\
    DDGR \citep{geo2023ddgr} & 93.5 $\pm$ 0.1 &37.0 $\pm$ 0.1 & 7.38 $\pm$ 0.01 \\
    \textbf{A\&A GAN (ours)} &\textbf{95.7} $\pm$ 0.1 & \textbf{51.1 $\pm$ 0.6}  & \textbf{16.6 $\pm$ 0.3} \\
\end{tabular}
    \caption{Comparison of our approach to other state-of-the-art methods in terms of accuracy after the final task in class-incremental setup. Results from our reproduction from provided code or~\cite{prabhu2020gdumb}. *For BIR on CIFAR10 and CIFAR100 we freeze the encoder after the first task.}
    \label{tab:comparison}
\end{table}

\section{Conclusion}

In this work, we introduce \ours{} -- a new method for continual learning based on continual alignment of the generative model's latent representations. 
We observe that our approach outperforms recent state-of-the-art methods in continual generative modeling in both standard class-incremental scenarios and our newly proposed knowledge-consolidation approach. 
To our knowledge, this is the first work that experimentally shows that with continually growing data from even partially similar distribution, we can observe both forward and backward knowledge transfer, resulting in improved performance. 
On top of the continual generative modeling task, we present a proof-of-concept solution using aligned representations from a generative model for the downstream classification task.

\section*{Acknowledgments}
This research was funded by the Polish National Science Centre under agreement no. 2020/39/B/ST6/01511, as well as by the Young PW project granted by Warsaw University of Technology within the Excellence Initiative: Research University (IDUB) program. The computing infrastructure was supported in part by PL-Grid Infrastructure grant nr PLG/2022/016058 and PLG/2023/016278.
 \bibliographystyle{elsarticle-num} 
 \bibliography{cas-refs}

\appendix

\include{supplementary}

\end{document}

%% file: defs.tex
\def\Eparam{\theta} 
\def\Dparam{\omega}

\def\comment#1{}

\def\eqref#1{(\ref{#1})}
\def\Beq#1\Eeq{\begin{equation}#1\end{equation}}
\def\Beqo#1\Eeqo{\begin{equation*}#1\end{equation*}}
\def\Beqs#1\Eeqs{\begin{align}#1\end{align}}
\def\Beqso#1\Eeqso{\begin{align*}#1\end{align*}}

%% file: supplementary.tex
\appendix
\section*{Appendix}
This is a supplementary material complementing our submission in which we present extended visualizations of the experiments with \ours{}, as well as the implementation details for our models. Finally, we show additional generations sampled from our generative model trained in the continual learning scenarios with the complex datasets such as combined MNIST $\rightarrow$ FashionMNIST and CelebA. 
\section{Discussion on the usage of task index in generative continual learning}
Access to the task code in continual learning of discriminative models simplifies the problem. It is mostly used when making crucial decisions, such as selecting the relevant part of the model for inference or the final classification decision. In such cases, a need for task code greatly undermines the universality of a solution.

Contrary to the {\it discriminative} models, in {\it generative} case, conditioning generation on task index does not influence or simplify the evaluation setting. The goal of a continually learned generative model is to generate an instance modeled on examples from {\it any} of the previous batches. Hence, to use a continually learned generative model in practice, we can randomly sample a task index (provided that it is lower than the total number of seen tasks) the same way we randomly sample input noise to the decoder or generator. 
In fact, training generative models with task index significantly simplify a class incremental scenario, in which data distributions from separate tasks -- with different classes are easily distinguishable from each other. In such cases, the task index serves as an additional conditioning input, imperceptibly leading to the conditional generative model. This limits the universality of the proposed generative continual learning method.

\begin{table}[!t]
\centering
  \resizebox{\textwidth}{!}{
\begin{tabular}{l|l|l|l||l|l}
\hline
\textit{dataset} & \textit{scenario} & \textit{n. tasks} & \textit{network} & \begin{tabular}[c]{@{}l@{}}Accuracy\\(VAE)\end{tabular} & \begin{tabular}[c]{@{}l@{}}Accuracy\\(GAN)\end{tabular} \\ \hline
MNIST            & Class-Incremental & 5                 & MLP400           & 96.809       & 98.46        \\
MNIST            & Dirichlet $\alpha=1$ & 10             & MLP400           & 94.960       & 96.73        \\
FashionMNIST     & Class-Incremental & 5                 & MLP400           & 81.412       & 83.75        \\
FashionMNIST     & Dirichlet $\alpha=1$ & 10             & MLP400           & 80.734       & 82.21        \\
DoubleMNIST      & Class-Incremental & 10                & MLP400           & 67.485       & 70.07        \\
\end{tabular}
}
\caption{Accuracy after the final task in different scenarios. Both in VAE and GAN-based approaches, the final accuracy on the Dirichled-based scenario is lower than in the class incremental case.}
\label{tab:results}
\end{table}

\section{Models architectures and implementation details}
\label{app:model_arch}
In this section, we describe in detail the architectures of models used in our experiments. The same models and hyperparameters can be found in the codebase.

\subsection{Fully connected Variational Autoencoder}

For comparison with other methods, we propose a simple VAE architecture with 9 fully connected layers, which we use with simpler datasets: MNIST, Omniglot, FashionMNIST, CERN and combined datasets MNIST $\rightarrow$ FashionMNIST and FashionMNIST $\rightarrow$ MNIST.

In the encoder, we use three fully connected layers transforming input of 784 values through 512, 128 to 64 neurons. Afterward, we map encoded images into continuous and binary latent spaces. For MNIST, we use continuous latent space of size 8 and additional binary latent with size 4. For FashionMNIST and Omniglot, we extend it to 12 continuous and 4 binary values. 

The translator network takes three separate inputs: continuous encodings, binary encodings, and binary codes representing task numbers. We first process both binary inputs separately through two fully connected layers of 18 and 12 values for task codes and 8 and 12 neurons for binary encodings. Afterward, we concatenate those three inputs: continuous noise from the encoder and two preprocessed binary encodings into a vector of size 32 for MNIST and 36 for Omniglot and FashionMNIST. We further process these values through two fully connected layers of 192 and 384 neurons, which is the size of the second latent space $\mathcal{Z}$

Our decoder consists of 3 fully connected layers with 512, 1024, and final 784 values. In each hidden layer of the model (except for the outputs of the encoder and translator), we use a LeakyRelu activation and sigmoid for the final one.

\begin{figure*}[!htbp]
\centering
\subfigure[Original data]{\includegraphics[width=.3\linewidth]{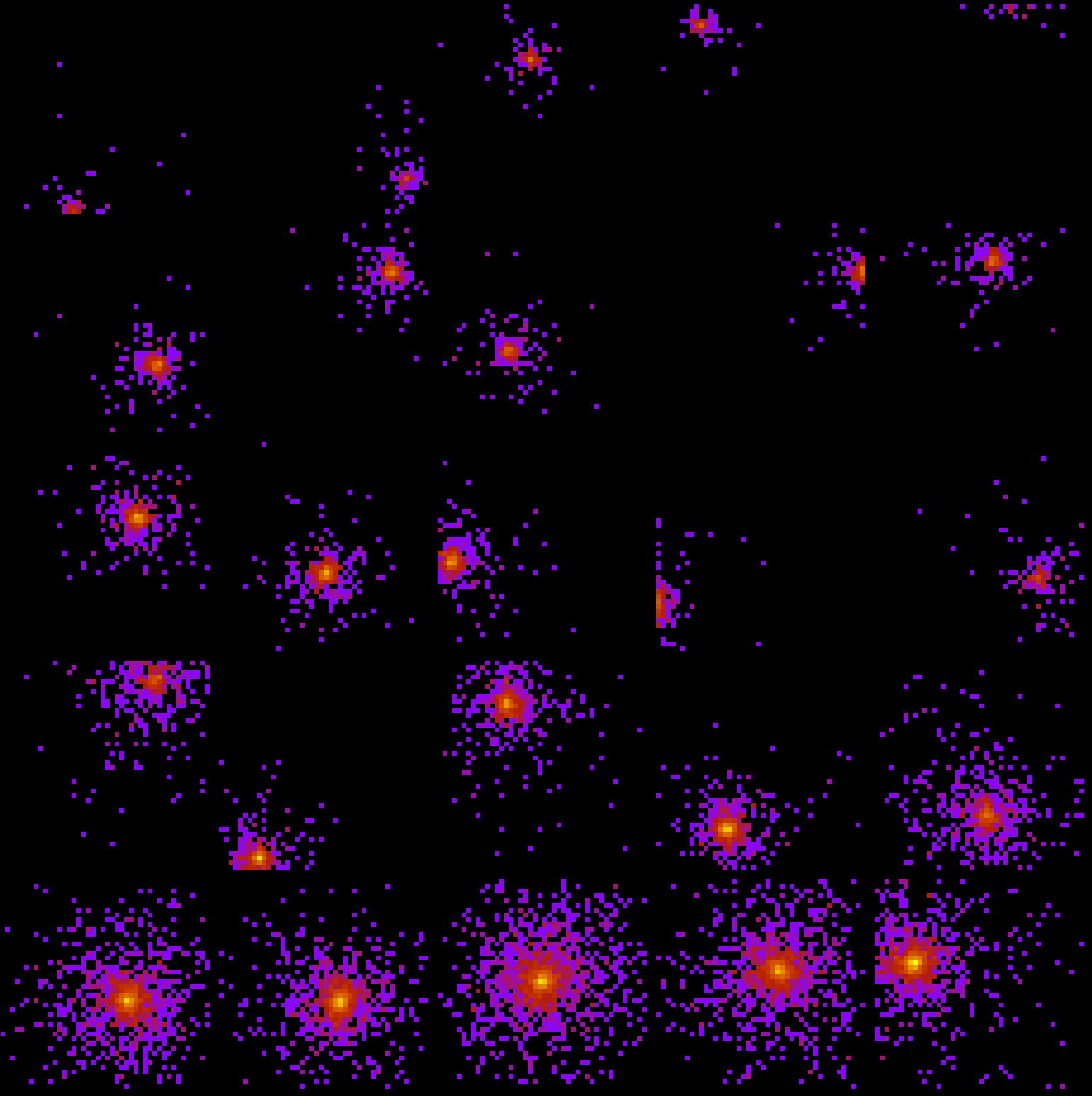}}
\subfigure[Generated simulations]{\includegraphics[width=.3\linewidth]{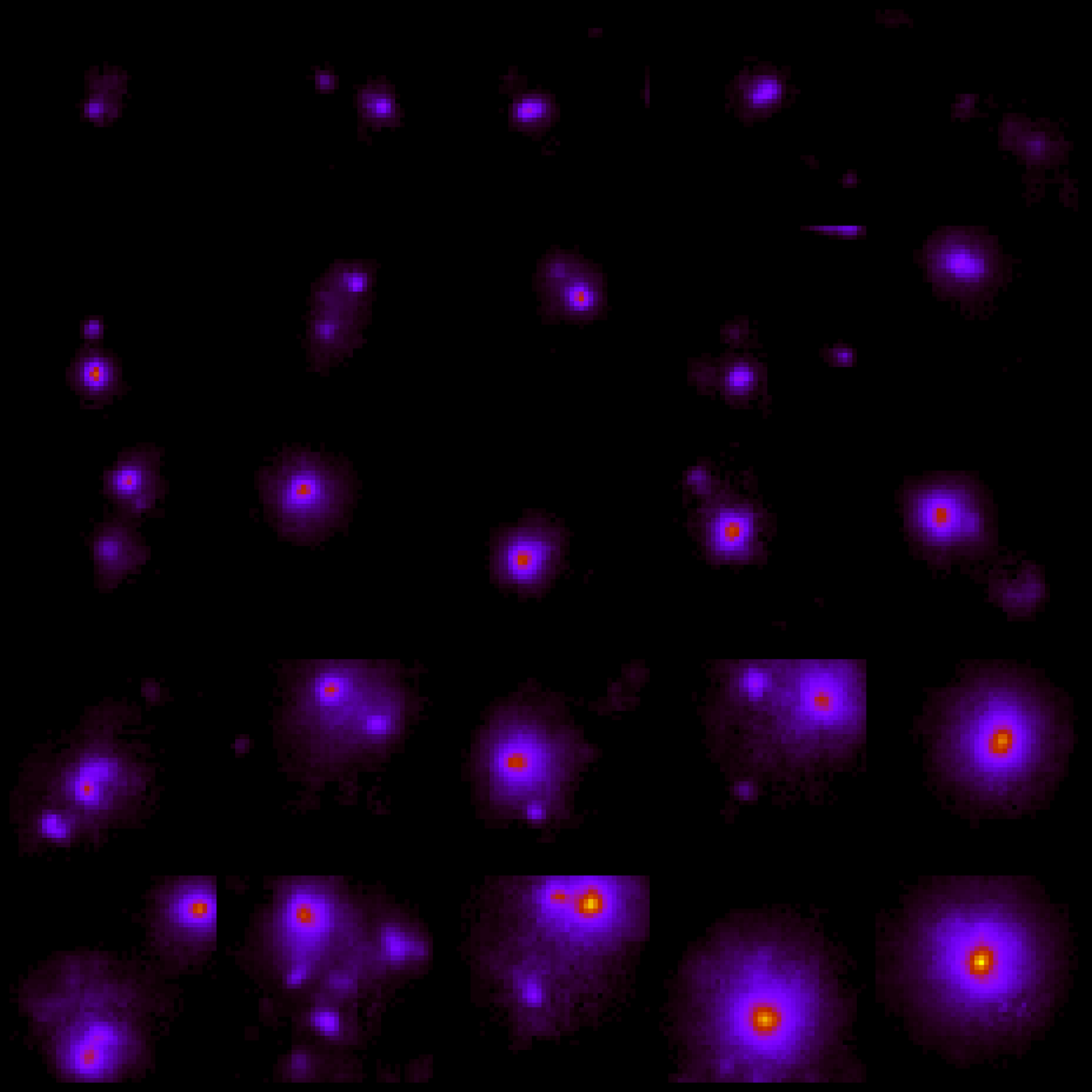}}
\caption{Original simulations for Zero Degree Calorimeter responses and generations from our \ours{} trained in the class incremental scenario on CERN dataset (\textbf{in logarithmic scale}). We split the original dataset into 5 tasks (each row of visualization) with increasing energy of input particle. This results in the continuously scaled size of the observed showers with partial overlapping between tasks. \ours{} well consolidates knowledge generating various outputs with full energy spectrum. Although because of the logarithmic scale generated examples seem blurred, this is of low importance because of the extremely low values of darker/purple pixels.}
\label{fig:cern_input_example} 
\end{figure*}

\subsection{Convolutional VAE}
Our \ours{} is not restricted to any particular architecture. Therefore, we also include experiments with a convolutional version of our model. In this setup, for the encoder, we use 3 convolutional layers with 32, filers each of $4\times4$ kernel size and $2\times2$ stride. After that, we encode the resulting feature map of 288 features into the latent spaces of the same size as the fully connected model.
For the translator network, we use a similar multilayered perceptron as in the fully connected model; however, we extend the dimensionality of latent space $\mathcal{Z}$ to 512.

In the decoder, we use one fully connected layer that maps the translator's output with 512 values into 2048 features. Those are propagated through 3 transposed convolution layers with 128, 64, and 32 filters of $4\times4$ kernel size and $2\times2$, $2\times2$, and $1\times1$ stride. The final transposed convolution layer translates filters into the final output with $4\times4$ kernel. 

For the CelebA dataset, we extend our convolutional model. In the encoder, we use four convolutional layers with 50, 100, and 200 filters with $5\times5$ kernel size, followed by fully connected layer mapping 1800 features through the layer of 200 neurons into the latent space of 32 neurons and binary latent space of 8 neurons. 
In the translator, we extend the fully connected combined layers into 800 and 1600 features which is a dimensionality of latent space $Z$. Our decoder decodes 1600 features from latent space through 3 transposed convolution layers with 400, 200, and 100 filters into the final output with 3 channels.

As in the fully connected model, we use LeakyReLU activations and additional batch normalization after each convolutional layer.

\subsection{Convolutional WGAN}
For a fair comparison, we use a WGAN architecture as similar to the convolutional VAE architecture as possible. To that end, we modify the VAE's encoder and decoder neural networks so they can be used as WGAN's critic and generator, respectively. 

Changes that have to be applied to VAE encoder architecture in order to construct a valid WGAN model: 
\begin{enumerate}
    \item We remove batch normalization layers, as they cannot be applied to WGAN's critic model since we penalize the norm of the critic's gradient wrt. each input independently. These layers are replaced with layer normalization,
    \item WGAN's critic model, as a classifier, outputs a single scalar.
\end{enumerate}

Moreover, we use exactly the same translator architecture, but we change the dimensionality of latent space  $\mathcal{Z}$ to 100. We also use the hyperbolic tangent activation function at the end of the generator model to obtain images in the range [-1, 1]. Other than that, on benchmarks where both GAN and VAE approaches are evaluated, GAN models have exactly the same hyperparameters as VAE models.

For the CIFAR10 and CIFAR100 datasets, we extend the architecture of GAN. In the critic model, we use 5 convolutional layers with respectively with 128, 256, 512, 1024 and 1024 filters. The first 4 layers have kernels of sizes 4 $\times$ 4 and strides 2 $\times$ 2. Only the last layer has 3 $\times$ 3 kernel and 1 $\times$ 1 stride. Each layer has a single padding and does not add learnable bias. On the other hand, in the generator model, we use 4 transposed convolution layers with 1024, 512, 256 and 128 filters. The last transposed convolution layer generates final output with 3 channels. The first 4 layers have kernels of sizes 4 $\times$ 4 and strides 2 $\times$ 2, only the last layer has 5 $\times$ 5 kernel and 1 $\times$ 1 stride along with 2 $\times$ 2 padding.

\subsection{VAE Training hyperparameters}
We train our VAE models with the Adam optimizer, learning rate $0.001$, and exponential scheduler with a scheduler rate equal to $0.98$. In our experiments, we train our model for 70 epochs of local training and 140 epochs of global training, with 5 epochs of shared knowledge discovery. We combine each mini-batch of original data examples with generations from previous tasks reaching up to mini\_batch\_size$\times$num\_tasks$\times$0.5 samples per mini-batch.
 
For the splits according to the Dirichlet distribution, we substitute target generations with cosine similarity greater than $0.95$. For the class incremental scenario, we set this parameter to 1. Nevertheless, our experiments indicate that lowering this value to $0.9$ does not influence the model's performance.

\subsection{GAN Training hyperparameters}
We train our GAN models with the Adam optimizer with $\beta_1$ equal to 0 and $\beta_2$ equal to 0.9. In the local training phase, we use a learning rate equal to 0.0002 for both the critic and generator models. In the global training phase, we set the learning rate to 0.001. In both phases, we use exponential learning rate schedulers with a rate equal to 0.99 with batch size 64. We train local models for 120 epochs and a global generator for 200 epochs, with 5 epochs of shared knowledge discovery, like in VAE's approach. We optimize the generator model once per 5 critic's steps. We use $\lambda$ equal to 10 in calculating the gradient penalty.  

We use the same set of hyperparameters for every benchmark that we evaluate on.

\subsection{Details of classification setup}
In all of our experiments, we use the exact same architectures as the GDumb \cite{prabhu2020gdumb} method as feature extractors in order to fairly compare our approach to other methods presented in this work.

For simple datasets such as MNIST, FashionMNIST, and DoubleMNIST, as a feature extractor, we use simple MLP with 2 hidden layers with 400 neurons in each layer. For the CIFAR10 dataset, we use ResNet18 and Resnet32 for CIFAR100. For each dataset that we evaluate, we use a very simple neural network with a single hidden layer that has 100 neurons as a classifier.

In feature extractor training we use an Adam optimizer with a learning rate equal to 0.001 and an exponential scheduler with a rate of 0.99. We set the number of epochs to 100 and the batch size to 256.

In classifier training, we use the exact same setup as in the training of the feature extractor, except that we train the classifier only for 20 epochs.

\section{Real-life CERN dataset}
\label{app:cern}
In this work, we evaluate different continual learning generative methods with real-life examples of particle collision simulation datasets. For that end, we use data introduced in~\cite{deja2020end} that consists of 117~817 Zero Degree Calorimeter responses to colliding particles, calculated with the full GEANT4~\cite{incerti2018geant4} simulation tool. Each simulation starts with a single particle with given properties (such as momenta, type, or energy) propagated through the detector with a simulation tool that calculates the interaction of a particle with the detector's matter. In the case of calorimeters, the final output of those interactions is the total energy deposited in the calorimeter's fibers. In the case of the Zero Degree Calorimeter at ALICE, those fibers are arranged in a grid with $44 \times 44$ size. To simulate a continual learning scenario, we divided input data into 5 tasks according to the input particle's energy as presented in Fig~\ref{fig:cern_input_example}. Such split simulates changing conditions inside the LHC, where the energy of collided beams changes between different periods of data gathering.

\section{Two latents Variational Autoencoder}
In the global part of our training, we rely on the regularization of VAE's latent space.
In practice, when encoding examples from distinct classes into the same latent space of VAE, we can observe that some latent variables are used to distinguish encoded class, and therefore they do not follow the desired continuous distribution as observed by ~\cite{tomczak2018vae} and~\cite{mathieu2019disentangling}. The extended experimental analysis of this phenomenon can be found in the supplementary material.

Therefore, in this work, we propose a simple disentanglement method with an additional binary latent space that addresses this problem, similar to the one introduced in~\cite{ramapuram2020lifelongvae}.
To that end, we train our encoder to encode input data characteristics into a set of continuous variables $\mu_c$ and binary variables $\mu_b$, which are used to sample vectors $\lambda_c$ and $\lambda_b$ that together form $\lambda$ -- the input to the translator model. 
For the continuous variables, we follow the reparametrization trick introduced by~\cite{kingma2014autoencoding}. To sample vector $\lambda_c$, we train our encoder to generate two vectors: means $\mu_m$ and standard deviations $\mu_{\sigma}.$ 
Those vectors are used as parameters of Normal distribution from which we sample  $\lambda_c \sim \mathcal{N}(\mu_m,\text{diag}(\mu^2_\sigma))$. For binary variables, we introduce a similar procedure based on the Gumbel softmax by~\cite{jang2016categorical} approximation of sampling from Bernoulli distribution. 
Therefore, we train our encoder to produce probabilities $\mu_p$ 
with which we sample binary vectors $L_b \sim B(\mu_p)$. 
To allow generations of new data examples for continuous values, we regularize our encoder to generate vectors $\lambda_c$ from the standard normal distribution $\mathcal{N}(0,I)$ with a Kullback-Leibler divergence. For binary vectors $\lambda_b$, during inference, we approximate probabilities $\mu_p'$ with the average of probabilities $\mu_p$ for all of the examples in the train set. We calculate $\mu_p'$ during the last epoch of the local training.
Therefore, to generate new data examples we sample random continuous variables $\lambda_c \sim \mathcal{N}(0,I)$ and binary variables $\lambda_b \sim B(\mu_p')$ and propagate them through the translator and global decoder.

\section{Analysis of binary latent space}
\begin{figure}
	\centering
	\includegraphics[width=0.99\linewidth]{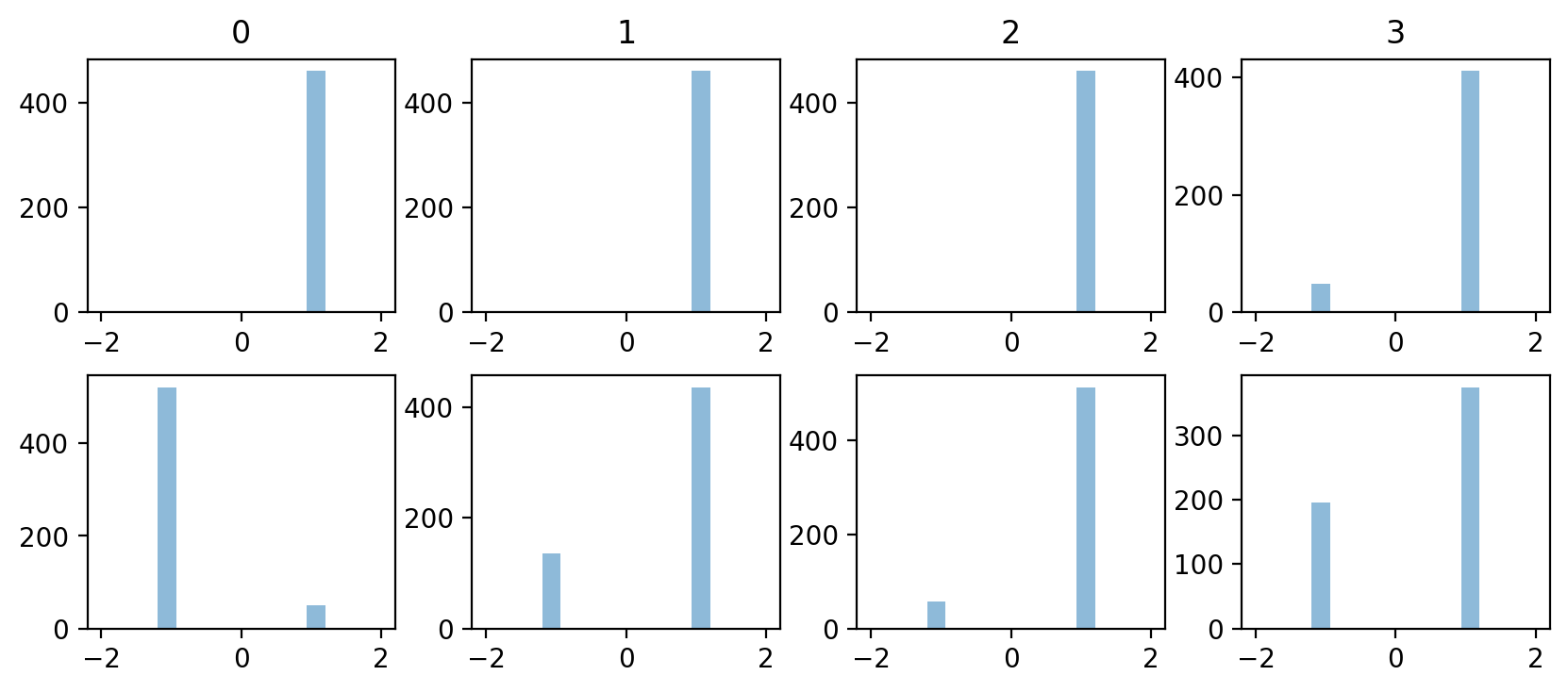}
	\caption{Binary latent space distribution of Variational Autoencoder. Sampled values for examples from encodings of class zero (top) and one (bottom). Additional binary latent space allows for simpler class separation mostly through the first binary value for which all of the zero examples are encoded with different values than for ones.}
	\label{fig:bin_latent_bin}
\end{figure}

\begin{figure}[tbh]
	\centering
	\includegraphics[width=0.45\linewidth]{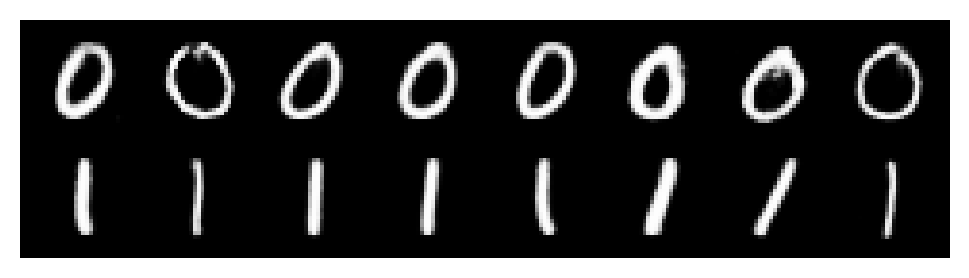}
	\caption{Examples of generations from Variational Autoencoder with binary latent space, for the same random continuous noise (per column) but opposite values for a first binary variable. As visible, our model disentangles classes through binary latent space while continuous values are still used to encode inter-class continuous features such as thickness or rotation.}
	\label{fig:gen_bin}
\end{figure}

\begin{figure}[tbh]
	\centering
	\includegraphics[width=0.45\linewidth]{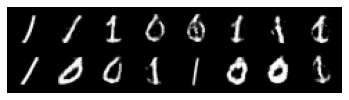}
	\caption{Examples of generations from Variational Autoencoder with no binary latent space, with variables 1 and 2 set to 0. Since the model uses those variables for class separation, resulting generations with sampled values around $0$ are between two classes.}
	\label{fig:gen_no_bin}
\end{figure}

\begin{figure*}[tbh]
	\centering
	\includegraphics[width=0.9\linewidth]{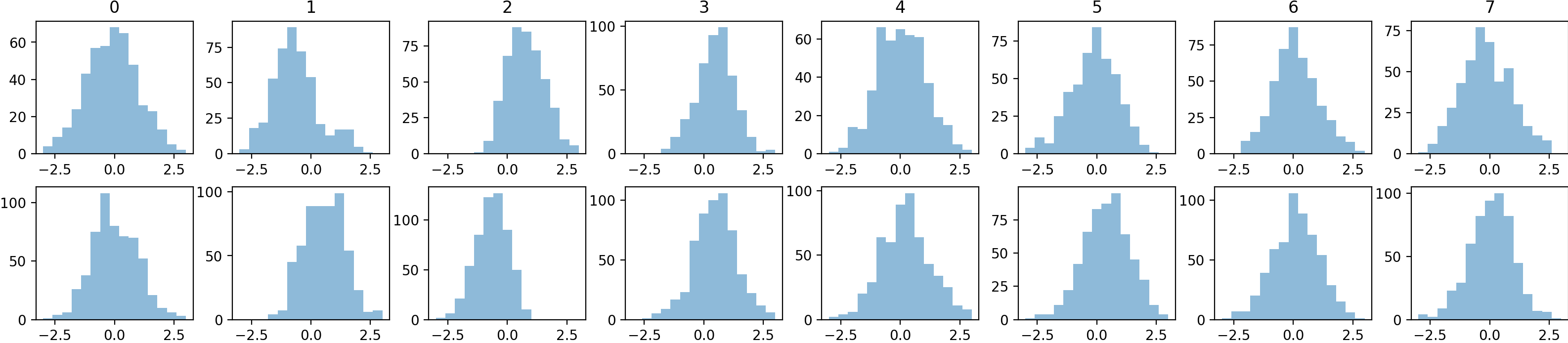}
	\caption{Latent space distribution of Variational Autoencoder trained with two separate classes. Noise embeddings for examples from class zero (top) and one (bottom). Two variables (1 and 2), do not follow standard normal distribution, but are used to differentiate examples from different classes.}
	\label{fig:latent_no_bin}
\end{figure*}

\begin{figure*}[tbh]
	\centering
	\includegraphics[width=0.99\linewidth]{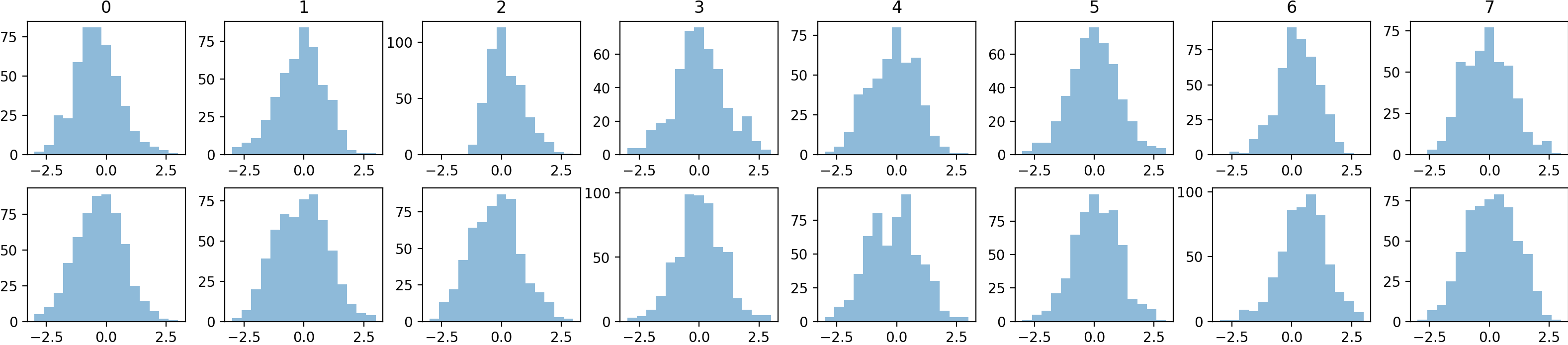}
	\caption{Latent space distribution of Variational Autoencoder with additional binary latent space trained with two separate classes. Noise embeddings for examples from class zero (top) and one (bottom). Thanks to the additional binary latent space, continual variables are better aligned to the standard normal distribution.}
	\label{fig:latent_bin}
\end{figure*}

When training Variational Autoencoder with complex data distributions such as a combination of several classes, we can observe that some of the variables in the latent space do not follow the desired distribution (e.g. $N(0,1)$), but instead, they are used to separate latent space into different parts. In this section, we explain this behavior on the basis of a simple example, with VAE trained on two classes from the MNIST dataset: zeros and ones. For that purpose, we analyze the latent space of the model. In Fig~\ref{fig:latent_no_bin}, we present a distribution of continuous variables when encoding examples from separate classes. As visible, two variables (1 and 2) do not follow the standard normal distribution to which they were regularized. Instead, they are used to differentiate examples from different classes. Therefore, for certain sampled values, e.g., with variable 2 around 0, the model generates examples that are in between two classes as presented in Fig.~\ref{fig:gen_no_bin}. With generative replay, this problem is even more profound since the rehearsal procedure leads to error accumulation.

In this work, we propose a simple disentanglement mechanism. In the process of data encoding, we use an additional binary latent space to which the encoder can map categorical features of the input data, such as distinctive classes. This simplifies encoding in standard continuous latent space in which our model does not have to separate examples from different parts of the original distribution. Compared with standard VAE, we extended the previous model with an additional binary latent space of four binary variables. After training with the same subset of the MNIST dataset of zeros and ones, we observe that the model encodes information about classes in the first binary variable as presented in Fig.~\ref{fig:bin_latent_bin}. With such binary codes, our autoencoder does not have to separate classes in the continuous latent space, which leads to better alignment to the normal distribution as presented in Fig.~\ref{fig:latent_bin}. In Fig.~\ref{fig:gen_bin}, we show sampled generations from our disentangled representation with two latent spaces. Samples in the same column share the same continuous noise, while those in the same row have the same binary vector.
Visualization indicates that continuous features, such as digit width or rotation, are shared between different binary features (column-wise), while for the same binary features (row-wise), we have only examples from the same class.

\begin{figure*}[t!]
\centering
\subfigure[MNIST~$\rightarrow$~FashionMNIST]{\includegraphics[width=.4\linewidth]{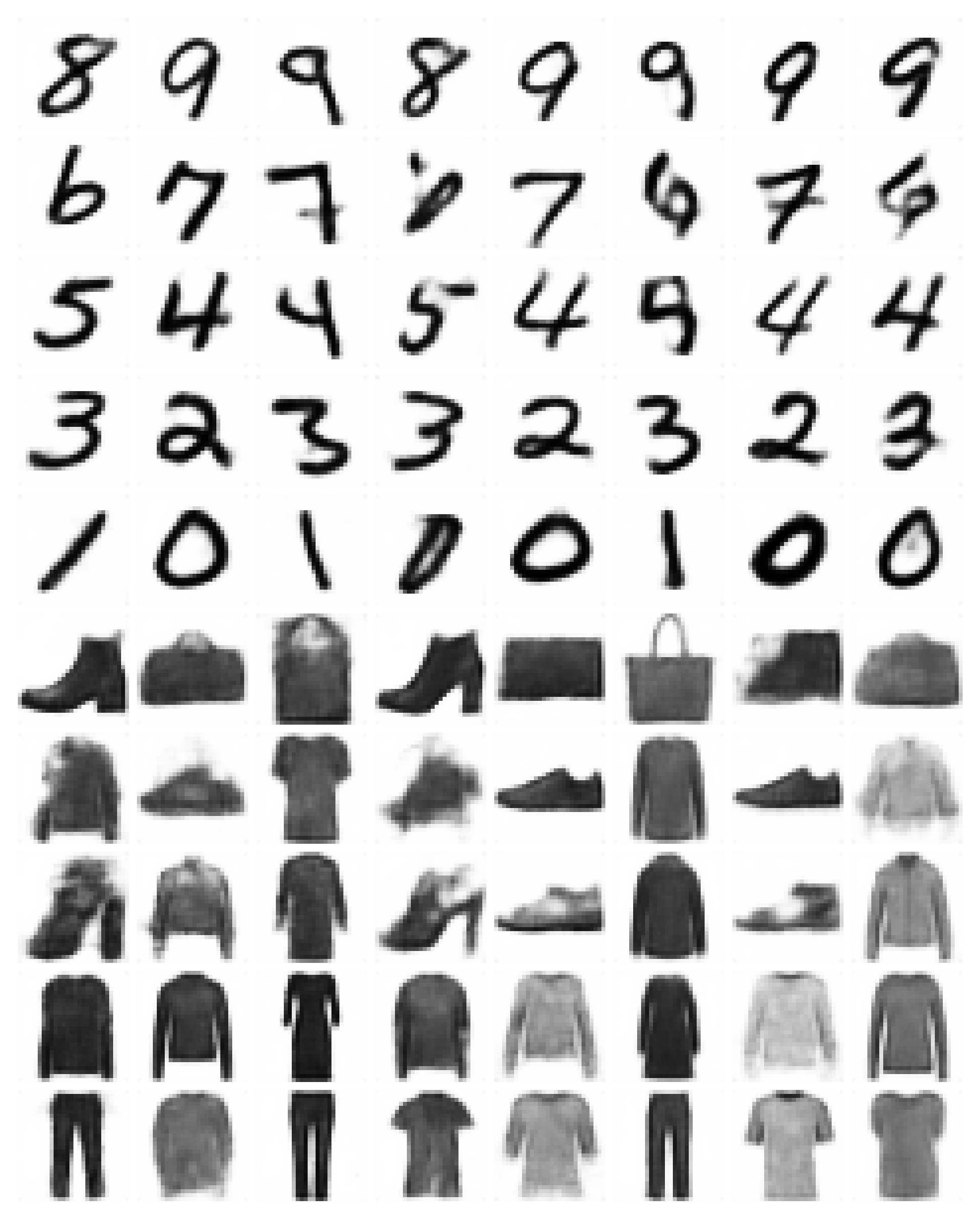}}
\subfigure[FashionMNIST~$\rightarrow$~MNIST]{\includegraphics[width=.4\linewidth]{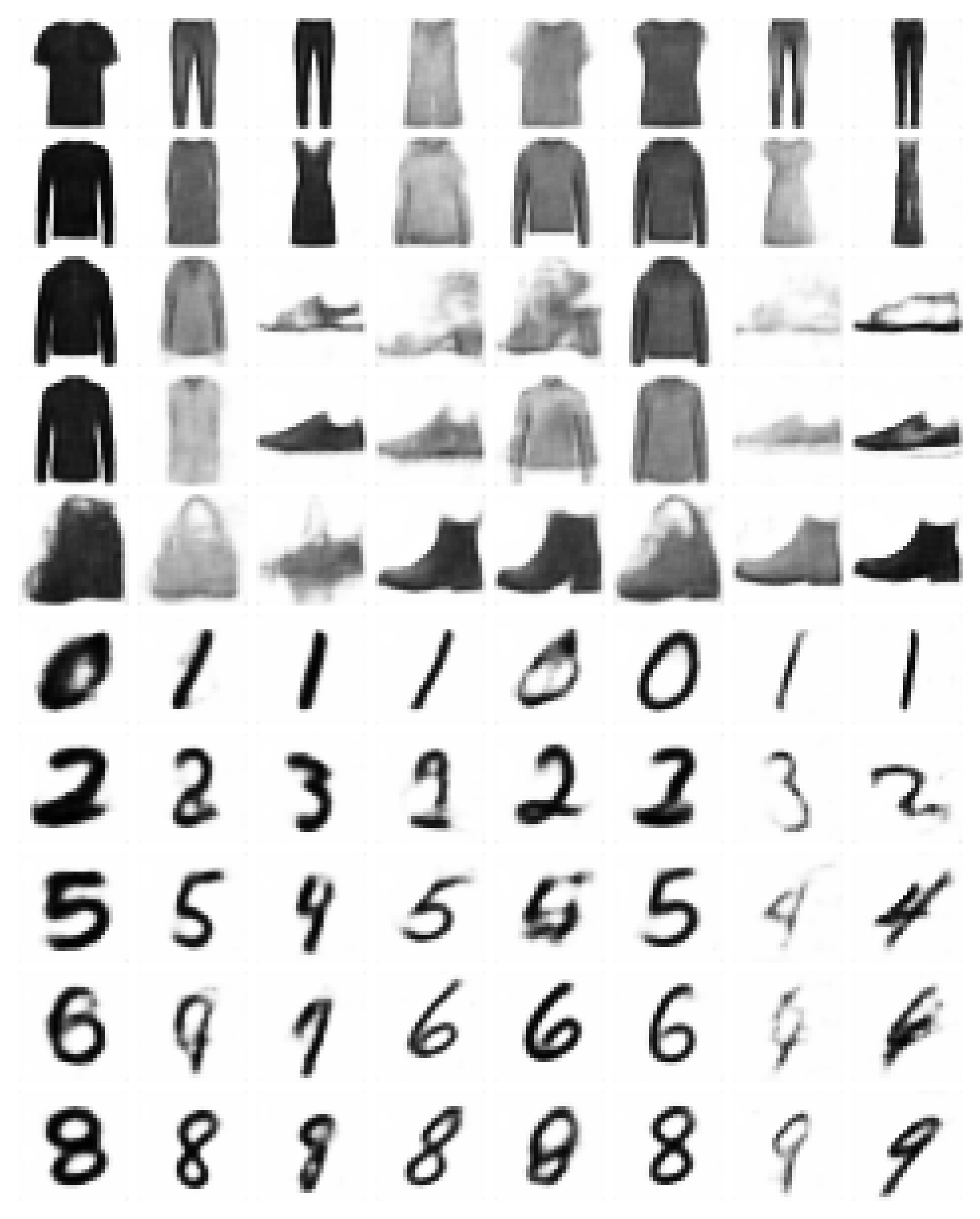}}
\caption{Images generated by our \ours{} trained in the class incremental scenario on combined datasets MNIST~$\rightarrow$~FashionMNIST (left) and FashionMNIST~$\rightarrow$~MNIST (right). We generate images with the same continuous noise per column. Thanks to the proposed band arrangement procedure, we can see that even when trained on drastically different distributions, our model adjusts data encodings from various tasks so that they share some common features. For example, in the first column of generations from FashionMNIST~$\rightarrow$~MNIST, we can observe how generations of thick black clothes correspond to the firm and bold instances of handwritten digits.}
\label{fig:generations_double} 
\end{figure*}

\begin{figure}[!b]
\vspace{3cm}
\centering
\includegraphics[width=0.45\linewidth]{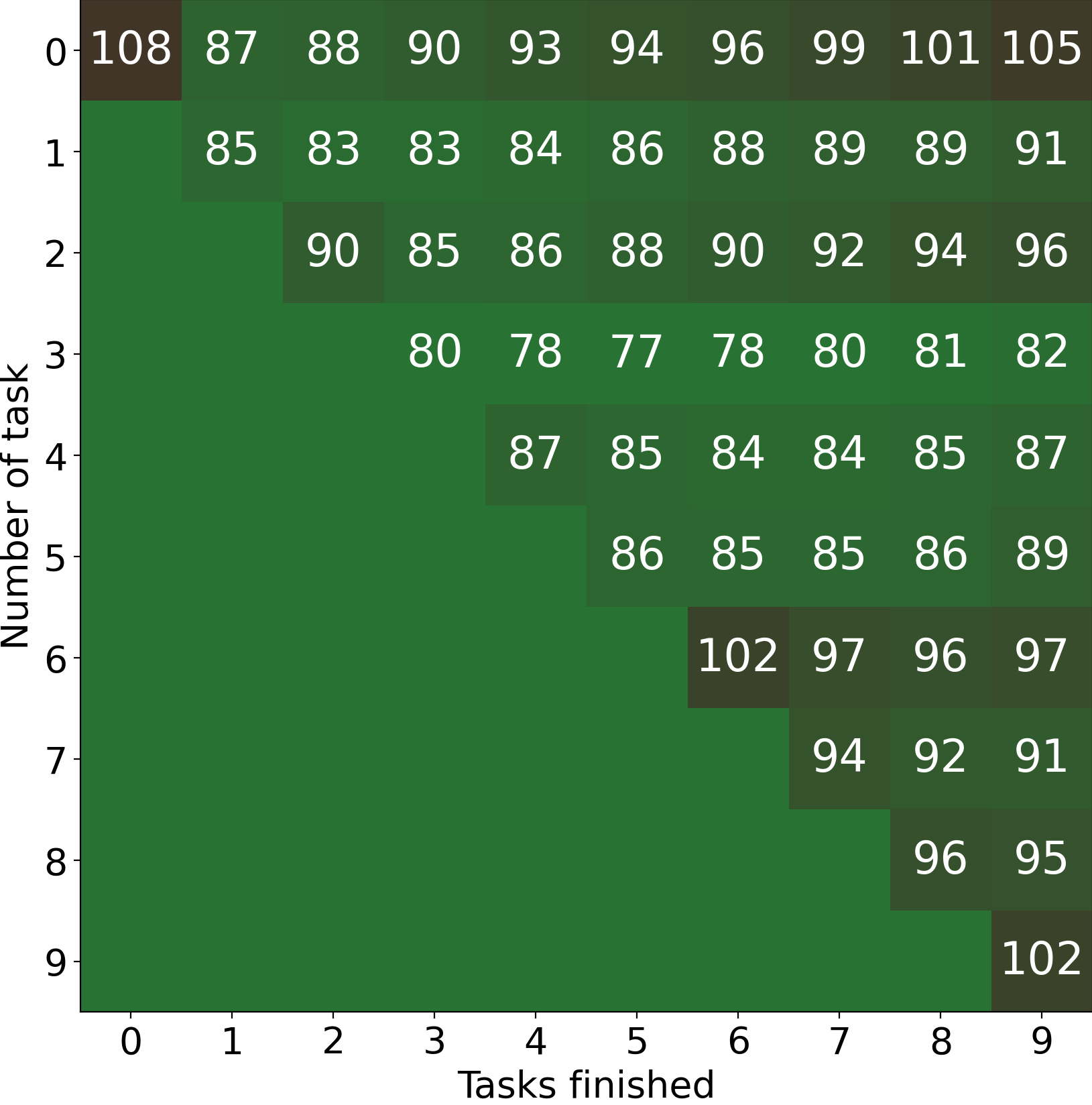}
\caption{ FID$\downarrow$ of generations from a given task of the CelebA dataset, after retraining with a number of following tasks for Dirichlet $\alpha=1$ scenario. Our \ours{} well consolidates knowledge with forward and backward knowledge transfer to generations from previous tasks when presented with new similar examples.\vspace{3cm}} 
\label{fig:celeba_example} 
\end{figure} 

\begin{figure}[tbh]
	\centering
	\includegraphics[width=0.7\linewidth]{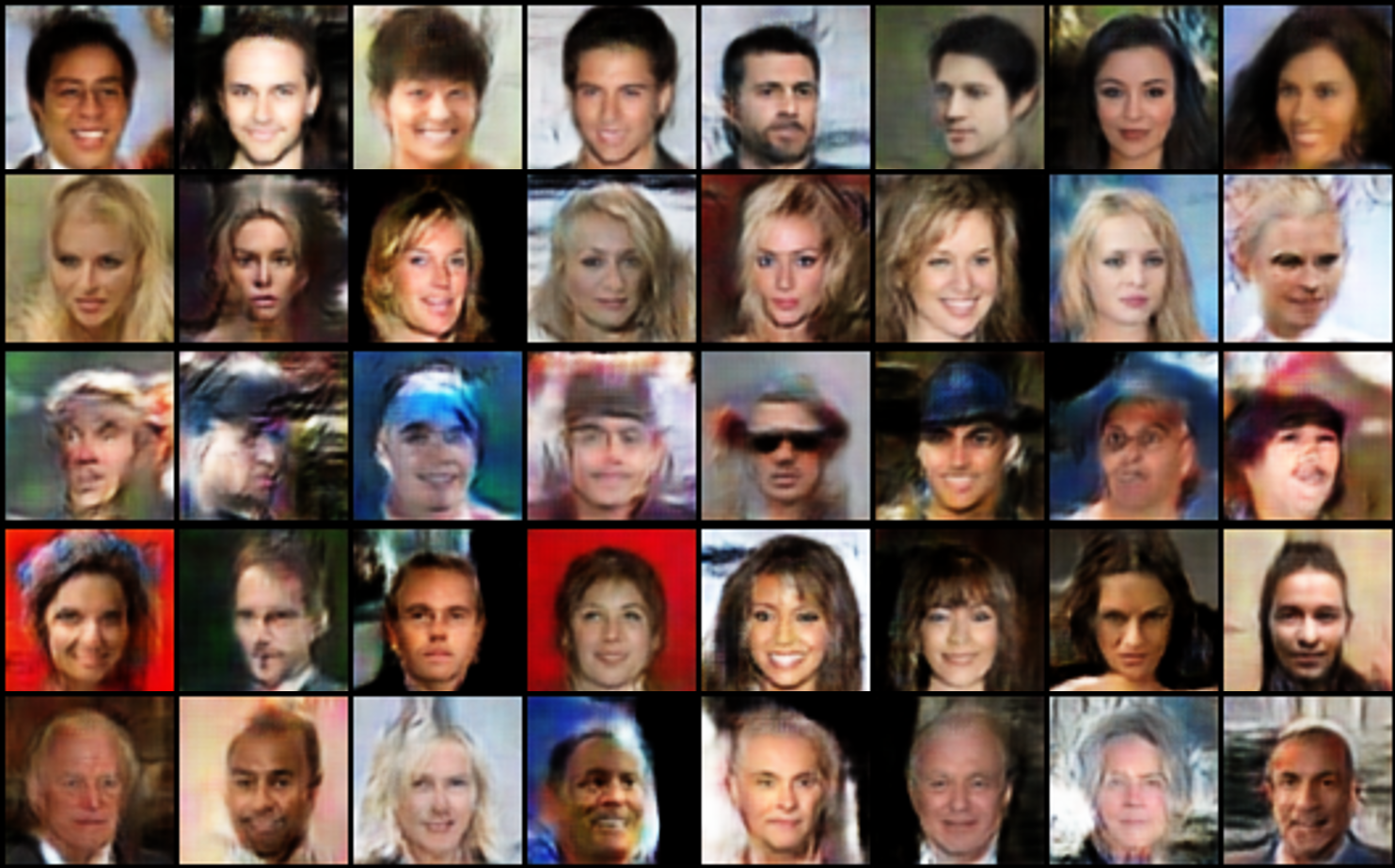}
 
	\caption{Images generated by \ours{} in the class incremental scenario for CelebA dataset. In the following tasks, we introduce images with different hair features. In the first task, we introduce photographs of people with black hair, followed by blondes, hats, and brown hair. In the final task, we trained the model with bald and white-haired people. In this visualization, we present samples with the same random continuous noise (per column) but different task indexes. We can observe that our \ours{} does not suffer from catastrophic forgetting.}
	\label{fig:geneations_celeba}
\end{figure}

\begin{figure}[tbh]
	\centering
	\includegraphics[width=0.7\linewidth]{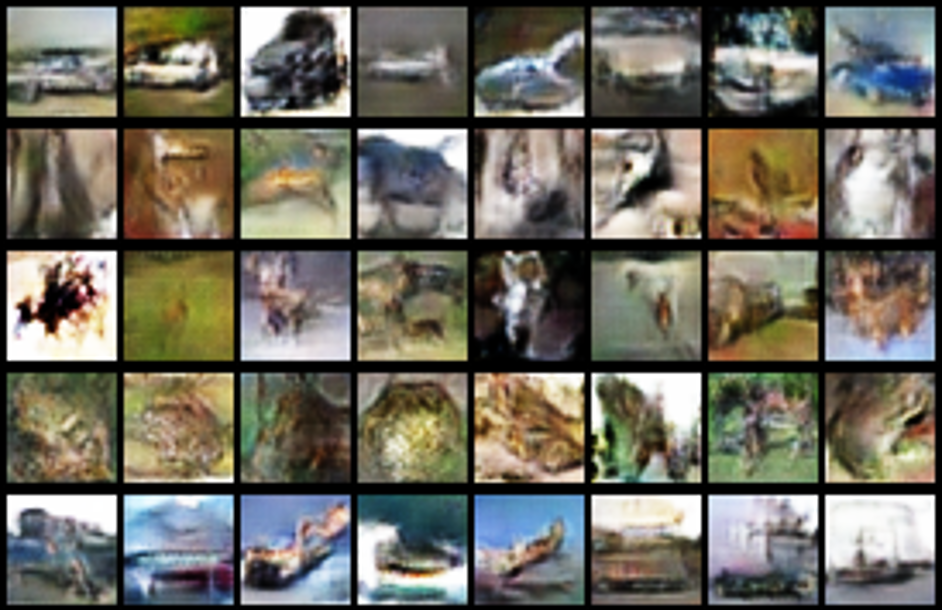}
 
	\caption{Images generated by ~\ours{} in the class incremental scenario from CIFAR10 dataset. Each row presents samples from the next task. We deduce from the generations' quality that this is a significantly more challenging setup than the previous one we evaluated. Nonetheless, the quality does not drop substantially compared to the upper bound.}
	\label{fig:generations_cifar}
\end{figure}

\section{Visualization of generated samples}
In this section, we present additional generations from \ours{}. Fig.~\ref{fig:generations_double} shows generations from combined datasets MNIST~$\rightarrow$~FashionMNIST and FashionMNIST~$\rightarrow$~MNIST. Our model does not suffer from catastrophic forgetting, so previous generations retain their good quality even when retrained with data samples from an entirely different dataset. Moreover, it is able to identify common features between datasets, such as the thickness of generated instances or their general shape. To visualize this behavior, we generate samples from the same instance of random continuous noise (column-wise) but conditioned on different task numbers.

In Fig.~\ref{fig:celeba_example}, we present one more example of how our knowledge consolidation works in practice on a standard benchmark. In most cases, the quality of new generations from the model retrained on top of the current global models is better than the previous one. Additionally, for some tasks, we can observe backward knowledge transfer in which training on the new task improves generations from the previous ones.


\section{CIFAR10 and CIFAR100 experiments}
\label{app:classification_results}
We evaluate our GAN-based approach on CIFAR10 and CIFAR100 datasets. From the results of upper bounds, presented in Tab.~\ref{tab:results_cifar}, we deduce that this setup is considerably more challenging than setups presented before. Nevertheless, our model is able to obtain satisfying results on the CIFAR10 dataset trained with class incremental scenarios. We observe worse results for the CIFAR100 dataset, where our model is not able to correctly learn the distribution of whole data from 20 separate portions, hence the large gap between the quality of samples for upper bound and class incremental scenarios.

\begin{table*}[!htbp]
  \centering
  \small
  \resizebox{\textwidth}{!}{
  \begin{tabular}{ccc|ccc|ccc||ccc} 
    \midrule
    Dataset & Scenario & Num. tasks  &FID$\downarrow$&Prec$\uparrow$&Rec$\uparrow$ \\
    \midrule
    CIFAR10 & Upper bound & 1 & 43 & 92 & 79 \\
    CIFAR10 & Class-Incremental & 5 & 73 & 86 & 65 \\
    CIFAR100 & Upper bound & 1 & 40 & 96 & 84 \\
    CIFAR100 & Class-Incremental & 20 & 144 & 77& 52 \\
    \bottomrule
  \end{tabular}
  }
  \caption{\label{tab:results_cifar} Average Fréchet Inception Distance (FID), distribution Precision (Prec), and Recall (Rec) for Multiband GAN approach trained on CIFAR datasets. We compare upper bounds to models trained with class incremental scenarios. Despite challenging setups, GAN models are able to consolidate knowledge from consecutive tasks.}
\end{table*}